\documentclass[11pt]{article}

\usepackage[utf8]{inputenc}
\usepackage[T1]{fontenc}
\usepackage[letterpaper,margin=1in]{geometry}
\usepackage{microtype}
\usepackage{amsmath,amssymb}
\usepackage{graphicx}
\usepackage{booktabs,longtable,multirow}
\usepackage{xcolor}
\usepackage{algorithm}
\usepackage{algorithmic}

\usepackage{listings}
\usepackage{tcolorbox}
\tcbuselibrary{listings, breakable}

\newtcblisting{promptlisting}{
    colback=gray!5,
    colframe=gray!50,
    boxrule=0.5pt,
    arc=2mm,
    left=5pt, right=5pt, top=5pt, bottom=5pt,
    listing only,
    breakable,
    listing options={
        basicstyle=\ttfamily\tiny,
        breaklines=true,
        columns=fullflexible,
        keepspaces=true,
    }
}

\usepackage[round,authoryear]{natbib}
\usepackage[hidelinks]{hyperref}
\hypersetup{
    pdftitle={Solver-Guided Reasoning for Mixed-Equilibrium Strategies},
    pdfauthor={Han Wang, Philippe Beardsell, Boning Li, Aaron Sasmita, Shuai Li, Hongyuan Zha, Baoxiang Wang}
}
\title{Solver-Guided Reasoning for Mixed-Equilibrium Strategies}

\author{%
Han Wang$^{1}$ \quad Philippe Beardsell$^{2}$ \quad Boning Li$^{3}$\\
Aaron Sasmita$^{4}$ \quad Shuai Li$^{1}$ \quad Hongyuan Zha$^{4}$ \quad
Baoxiang Wang$^{4,5}$\\[0.6em]
{\small $^{1}$Shanghai Jiao Tong University \quad $^{2}$GTO Wizard \quad
$^{3}$Tsinghua University}\\
{\small $^{4}$The Chinese University of Hong Kong, Shenzhen \quad
$^{5}$Vector Institute}\\[0.45em]
{\scriptsize \texttt{xwanghan@sjtu.edu.cn} \quad
\texttt{phil@gtowizard.com} \quad
\texttt{li-bn22@mails.tsinghua.edu.cn}}\\
{\scriptsize \texttt{121040033@link.cuhk.edu.cn} \quad
\texttt{shuaili8@sjtu.edu.cn} \quad
\texttt{zhahy@cuhk.edu.cn}}\\
{\scriptsize \texttt{bxiangwang@cuhk.edu.cn}}\\
}
\date{August 4, 2026}

\begin{document}

\maketitle

\begin{abstract}

Reasoning in large language models (LLMs) is often grounded in human text, human demonstrations, and human-generated rationales. For equilibrium reasoning in complex games, however, relying on human data can be suboptimal. In fact, human play is often guided by intuition and heuristics and can deviate substantially from game equilibrium. This discrepancy is amplified in games with mixed-strategy equilibria, where human data is heavily biased toward pure strategies. Consequently, conditioning LLMs on this data yields weak game strategies. To grant LLMs the reasoning capacity in games, in this work, we study how to elicit equilibrium play using solver output. We propose Mixed-Strategy Decision Tree (MDT), which articulates the silent optimality of the equilibrium into sparse strategic rules that both humans and LLMs could understand. Using solver output rather than human annotation allows us to extend the input to arbitrarily new states and continuations. We instantiate this study on No-Limit Texas Hold'em by querying a solver oracle for over \textbf{250 million mixed-strategy decisions}; MDT together with other techniques \textbf{reduces the $\ell_1$ distance to the equilibrium by $52.6\%$} across $8$ different LLM configurations. A Route-only ablation tests the incremental contribution of the shadow-based contrast, while complete River-endgame and Liar's Dice experiments evaluate strategic fidelity and portability beyond the original NLH communication setting.

\end{abstract}

\section{Introduction}\label{sec:intro}

Large language models are trained and evaluated with an extensive amount of human data, including demonstrations, solutions, rationales, and reasoning traces \cite{wei2022chain,achiam2023gpt}. For reasoning in complex games, however, learning from human data is fundamentally limited. On the one hand, human play and commentary are selective. Many human strategies are exploitative, largely deviating from game-theoretically optimal equilibrium strategies. Even players considered ``good'' by human standards utilize strategies that are mostly effective against weaker humans but underperform against AI solvers \cite{brown2019superhuman, silver2018general, silver2017mastering}. On the other hand, human data is mostly presented as pure strategies. It is intuitive for humans and LLMs to treat decisions as prediction tasks, attempting to figure out which single action yields the best outcome. But for games with imperfect information, the equilibrium strategy is often mixed. It is difficult to find such mixed strategies in human data, making it equally difficult for LLMs to acquire them.

Recent evaluations of LLMs on No-Limit Texas Hold'em (NLH) poker make this gap concrete. The GTO Wizard Benchmark reports that the strongest evaluated model, GPT-5.3 Extra High reasoning, performs substantially below its approximate equilibrium strategy, at roughly $-16 \pm 3.0$ bb/100 under its luck-adjusted evaluation \citep{gtowizardbenchmark2026}. LLMs also make fundamental errors in card representation and hand-strength evaluation, as well as strategic errors in range-level mixing and frequency allocation. PokerBench similarly observes that LLMs fail to predict the highest-frequency actions at certain decision points even after task-specific fine-tuning \citep{zhuang2025pokerbench}.
These failures are not simply a lack of poker vocabulary. LLMs can often discuss pot odds, blockers, equity, and bluffing, yet still miss the equilibrium logic that couples those concepts across hidden states and action frequencies.

This fundamental gap stems from a mismatch in the underlying objectives. LLMs are optimized for linguistic predictions and reasoning, whereas game solvers compute policies by minimizing exploitability across complex, hidden-state game trees \citep{zinkevich2007regret,moravvcik2017deepstack,brown2018superhuman}. Consequently, an LLM can produce fluent, conceptually accurate poker commentary while entirely failing to execute the precise frequency allocations required for equilibrium. Relying on human-generated examples or LLM self-rationalization cannot bridge this gap, as neither reliably captures full mixed-strategy policies \citep{lin2026how}. Conversely, while AI solvers naturally generate optimal mixed strategies, their outputs consist of raw numerical distributions rather than generalized, textual reasoning. To endow LLMs with game-theoretic optimality, we must translate these raw numbers into a linguistic format. We define this task as \emph{solver articulation}:

\begin{center}
\textit{How can we extract articulate, verifiable reasoning from \\
the silent optimality of solver-generated data?}
\end{center}

We study this problem in imperfect-information extensive-form games, using No-Limit Texas Hold'em (NLH) decision points where two players are left in the game. We introduce Mixed-Strategy Decision Tree (MDT), which converts solver-implied decision logic into an inspectable and readable form. MDT represents each decision point with its solver-derived public state and range-level and hand-level summaries. It uses sparse hierarchical routing to assign probability mass to pure-action leaf prototypes. The hierarchy avoids forcing all strategic interactions into a single dense mapping. In this way, coarse public and range conditions select a local strategic regime, while sparse node-level summaries expose the hand-specific boundary that changes the action mixture.

Motivated by counterfactual methods for game solving \citep{wachter2017counterfactual, agarwal2021neural}, we propose Scenario-Constrained Counterfactual Sampling (SCCS) that provides additional reasoning paths on top of MDT. SCCS selects shadow hands that share the same public context but exhibit clear solver-policy divergence, route to different MDT leaves, and differ along a salient summary. By isolating such contrastive pairs, SCCS exposes the local boundary at which a hand changes its role inside the mixed strategy. In this way, it converts an inspectable tree route into a transferable strategic statement.

Our method is applied to two-player NLH postflop game play with over 250 million decision points accessed. To ensure the quality of the articulation, we used one of the best available commercial poker solvers in the world. The solver enjoys a Nash Distance less than 0.3\% of the current pot and gives solutions to arbitrary spots in the game tree. The stream of data therefore spans all 1,755 NLH flops and their turn and river continuations. The obtained MDT is tested across 8 different LLM configurations, where the sparse rules by the MDT are given to LLMs before they are asked to reason the equilibrium strategy. The LLMs have their $\ell_1$ distance to the solver target reduced from $0.211$ to $0.100$. The argmax-action agreement, defined by the highest-probability action in each distribution, improves from $57.2\%$ to $76.1\%$.

Beyond the immediate improvements in game-theoretic reasoning, our framework sheds light on a fundamentally new regime in artificial intelligence: endowing LLMs with complex reasoning capabilities entirely through synthetic, AI-generated data. This paradigm is especially appealing in the current landscape, where the supply of high-quality human data is rapidly becoming depleted. By demonstrating that the implicit optimality of an AI solver can be systematically extracted and translated into readable linguistic rules, we provide a concrete pathway to bypass the human data bottleneck. As envisioned by \citet{silver2025welcome}, the future of artificial intelligence relies not on mimicking human demonstrations, but on learning directly from ground-truth interactions and solver-backed experience. \textbf{Our work represents an effort toward realizing this new era of experiences}.

\section{Related Work}

\paragraph{Solver-based game reasoning.}
CFR, subgame solving, and self-play search have enabled strong imperfect-information game agents, including superhuman poker systems \citep{zinkevich2007regret,brown2018superhuman,brown2019superhuman,moravvcik2017deepstack,brown2020combining}. These systems compute mixed policies and values, but their outputs are primarily numerical prescriptions rather than communicable reasoning.

\paragraph{Interpretable policy distillation.}
Prior work distills learned policies into trees, programs, or concept-based representations \citep{bastani2018verifiable,verma2018programmatically,frosst2017distilling,mcgrath2022acquisition}. Our setting differs because the target is a mixed equilibrium policy in an imperfect-information game, where local decisions depend on range-level coupling and action-frequency balance.

\paragraph{LLMs and poker reasoning.}
Recent poker benchmarks show that LLMs struggle with solver-level poker decisions despite fluent strategic language \citep{gupta2023chatgpt,zhuang2025pokerbench,gtowizardbenchmark2026}. Rather than training an LLM to play poker directly, we study how solver-derived mixed-strategy distinctions can be converted into contrastive rules that independent LLMs can use. Additional related work is discussed in Appendix~\ref{app:related_work}.

\section{Preliminaries}
\label{sec:preliminaries}

\paragraph{Imperfect-information extensive-form games.}
 No-Limit Texas Hold'em (NLH) is a zero-sum extensive-form game with imperfect information; Appendix~\ref{app:NLH_terms} summarizes the rule structure and domain vocabulary used throughout. Postflop decisions occur after public community cards are revealed: the flop is the betting round after three public cards, and the turn is the round after the fourth public card. An extensive-form game is defined by
$\mathcal{G}=(\mathcal{N},\mathcal{H},\mathcal{Z},\mathcal{A},P,u,\mathcal{I})$,
where $\mathcal{N}=\{1,2\}$ is the player set, $\mathcal{H}$ is the set of finite histories, $\mathcal{Z}\subset\mathcal{H}$ is the set of terminal histories, $\mathcal{A}(h)$ is the set of legal actions after a non-terminal history $h$, $P(h)\in\mathcal{N}\cup\{c\}$ specifies whether a player or chance acts at $h$, and $u_i(z)$ is player $i$'s payoff at terminal history $z$. The game is zero-sum, so $u_1(z)+u_2(z)=0$. In poker, a history contains public events such as betting actions and community cards, together with private cards dealt by chance.

Imperfect information is represented by information sets. For player $i$, $\mathcal{I}_i$ partitions the decision histories at which $i$ acts. Histories $h,h'\in I\in\mathcal{I}_i$ are indistinguishable to player $i$: they share the same public betting/card history and the same private hand for $i$, but may differ in the opponent's private hand. A behavioral strategy is therefore a distribution over actions at each information set,$\pi_i(\cdot\mid I)\in\Delta(\mathcal{A}(I))$. Throughout the paper, $h$ denotes a generic non-terminal decision history, while $z$ is reserved for terminal histories.

\paragraph{Nash equilibrium and GTO strategies.}
A strategy profile $\pi=(\pi_1,\pi_2)$ induces an expected utility $u_i(\pi)$ by integrating terminal utilities over chance outcomes and both players' randomized actions. A Nash equilibrium (NE) is a profile $\pi^*$ such that no player can improve by unilateral deviation \citep{nash1951noncooperative}:
\begin{equation}
u_i(\pi_i^*,\pi_{-i}^*) \geq u_i(\pi_i',\pi_{-i}^*) \qquad
\forall i\in\mathcal{N},\ \forall \pi_i'.
\end{equation}
In poker terminology, a Game-Theoretic Optimal (GTO) strategy is an approximate NE strategy computed by a solver. Modern poker solvers are commonly based on regret minimization, search, public-belief reasoning, and subgame solving \citep{zinkevich2007regret,bowling2015heads,moravvcik2017deepstack,brown2018superhuman}. Solver outputs are commonly reported as mixed action distributions together with action values or counterfactual values. For a decision history $h$, $\pi^*(\cdot\mid h)$ denotes the computed action distribution, and $Q^*(h,a)$ denotes the value associated with legal action $a\in\mathcal{A}(h)$. The notation $Q^*(I,a)$ is used when discussing the corresponding information-set formalism.

\paragraph{Public belief states and ranges.}
Solvers do not reason about a single fully observed state. Given a public state $s$ consisting of the public board and betting sequence, there are many private-card assignments consistent with what has been observed. The public belief state (PBS) \citep{brown2020combining}, also called a range representation in poker, is the conditional distribution over these private assignments:
$\beta_s(c_1,c_2)=\Pr(c_1,c_2\mid s)$, where $c_i$ denotes player $i$'s private hand. A player's range is the corresponding marginal distribution over that player's possible private hands. The solver's equilibrium policy can be viewed as a high-dimensional mapping from this public belief state and a particular private hand to a mixed action distribution: $f^*:\ (s,c_i,\beta_s)\mapsto \pi_i^*(\cdot\mid I_i(s,c_i))$. This mapping is high-dimensional because the action distribution for one private hand can depend on the public state, the player's range, the opponent's possible range, and continuation values induced by future play.

\paragraph{Indifference and mixed strategies in Nash equilibrium.}
Mixed strategies are often necessary in imperfect-information games because deterministic action patterns can reveal exploitable information. An observed action changes what an opponent can infer about hidden private states. If an action is associated too strongly with a narrow class of private states, the opponent may be able to respond profitably to that revealed structure. Equilibrium mixing helps prevent such profitable deviations by distributing probability mass across actions so that no player can improve unilaterally.

Indifference is a consequence of this equilibrium condition, not an independent assumption. If two actions are both used with positive probability at an information set, then neither can have strictly higher counterfactual value against the opponent's equilibrium strategy; otherwise probability could be shifted toward the better action. Thus supported actions have equal value up to approximation error, while unsupported actions have no higher value \citep{nash1951noncooperative,osborne1994course}. In games with perfect recall, behavioral strategies represent a player's randomized choices at information sets \citep{kuhn1953extensive}, and CFR-style algorithms approach equilibrium by minimizing counterfactual regret \citep{zinkevich2007regret}. Formally, for an information set $I$ and any action $a$ in the support of the equilibrium strategy,
\begin{equation}
Q_i^*(I,a)\approx V_i^*(I)
\qquad \text{for } a \text{ with } \pi_i^*(a\mid I)>0,
\end{equation}
with $Q_i^*(I,a)\le V_i^*(I)$ for actions outside the support.

\paragraph{Global dependence in imperfect-information games.}
The preceding definitions still leave the central difficulty: in imperfect-information games, a local decision generally cannot be interpreted as an isolated choice at a fully observed state. The value of a local strategy can depend on the belief over hidden states and on constraints imposed by the full-game strategy, rather than only on the visible public state. This is a standard obstacle in imperfect-information subgame solving: unlike perfect-information games, an optimal strategy for a reached subgame may depend on strategies in other, unreached parts of the game, so the subgame cannot be solved independently of the full-game strategy \citep{brown2017safe}.

\begin{figure}
    \centering
    \includegraphics[width=1\linewidth]{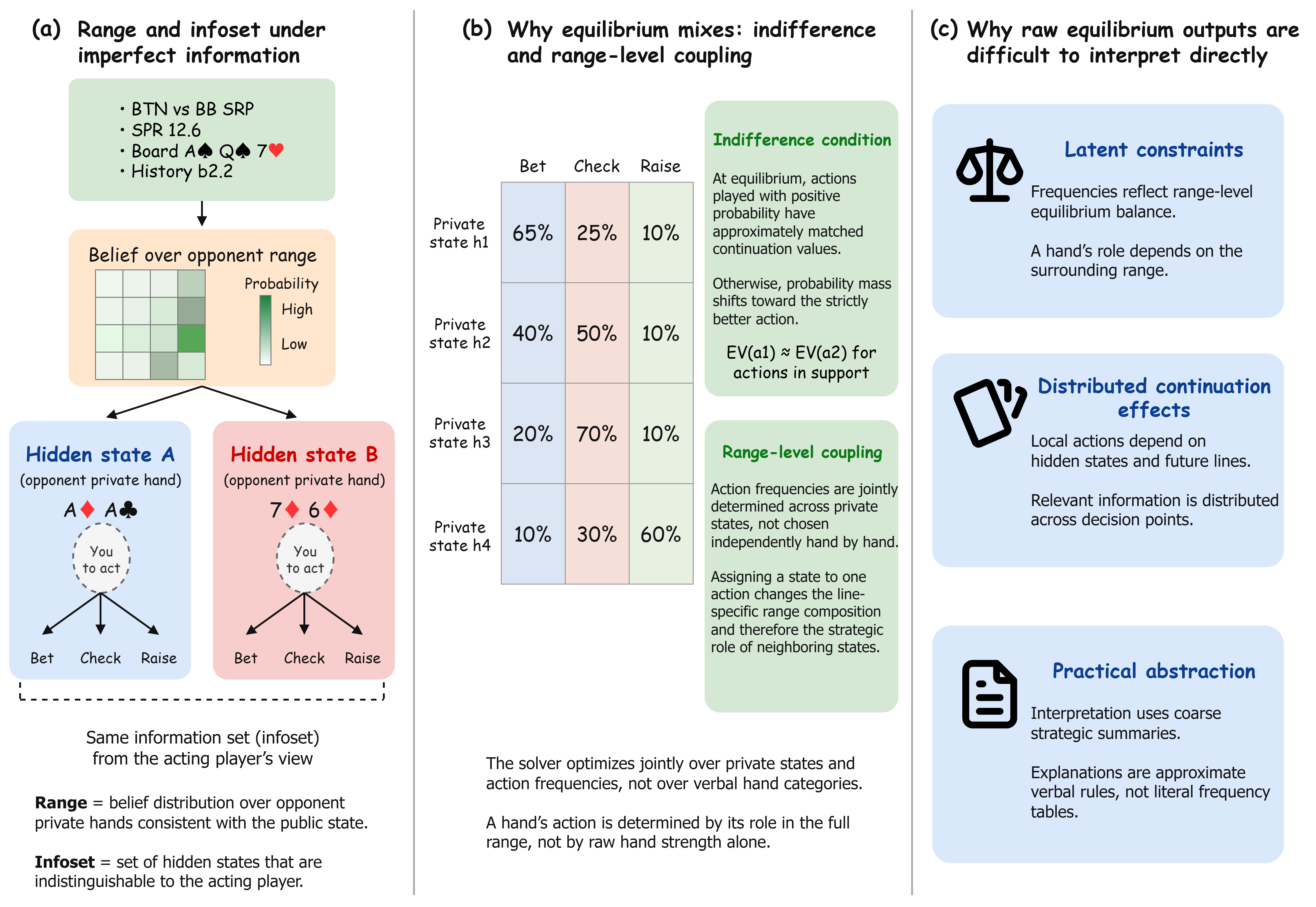}
    \caption{\textbf{Why equilibrium strategies in imperfect-information games are hard to interpret.} (a) The same public situation corresponds to many hidden private-card states inside one information set, so a visible action cannot be explained from public context alone. (b) Equilibrium mixing is constrained by indifference and belief-dependent continuation values, so action frequencies cannot be reduced to a single best action. (c) A solver table exposes the numerical policy and continuation values, but not the contrastive rule that makes strategically similar hands diverge; our goal is to compress this latent continuation logic into communicable strategic summaries.}
    \label{fig:challenge}
\end{figure}

\section{Why Equilibrium Mixing in Poker Is Difficult}
\label{sec:why-hard}

NLH is difficult for language models not only because the game tree is large, but because the strategic object to be learned is a \emph{mixed equilibrium policy} over imperfect-information states. The relevant target is not a single best action for a visible hand. It is a range-level allocation of action frequencies that remains hard to exploit after the opponent updates beliefs from the observed betting line. This distinction is central to modern poker AI: superhuman systems such as DeepStack, Libratus, and Pluribus rely on equilibrium-oriented search, self-play, abstraction, and subgame reasoning in hidden-information games rather than on human explanations alone \citep{moravvcik2017deepstack,brown2018superhuman,brown2019superhuman}. These results suggest that solver-generated equilibrium behavior is a more appropriate source of strategic targets than human verbal heuristics alone.

\paragraph{Publicly available poker discourse provides language, not equilibrium logic.}
Publicly available poker text is abundant but structurally mismatched to the object we need to learn. Forum posts, coaching examples, and hand histories are selective: they usually explain memorable or exploitative decisions, not the full support of a balanced range at an information set. This is consistent with recent LLM poker evaluations. PokerBench reports that strong pretrained LLMs substantially underperform on curated GTO decision spots, with GPT-4 reaching only $53.55\%$ accuracy before task-specific fine-tuning \citep{zhuang2025pokerbench}. Earlier work similarly finds that ChatGPT and GPT-4 can discuss starting-hand value, position, and GTO concepts while still failing to play game-theoretic optimal poker \citep{gupta2023chatgpt}. This gap does not mean that verbal heuristics are useless: conservative advice such as calling rather than raising in marginal aggressive nodes may reduce immediate losses. The limitation is that such advice is local, whereas equilibrium requires solver-level frequency allocation across the whole range. Our own pilot experiments with scraped Two Plus Two (2+2) forum text and SFT on an open Qwen model point in the same direction: next-token imitation can teach poker vocabulary and local heuristics, but it does not reliably recover basic range-level decision logic. We therefore treat publicly available poker discourse as a useful source of terminology and surface-level heuristics, but not as ground-truth evidence for equilibrium decision logic. Appendix~\ref{app:forum_sft_pilot} provides additional details.

\paragraph{Equilibrium mixing is functional, not incidental.}
In imperfect-information games, randomization is not noise around an underlying pure decision. It is a mechanism for controlling information leakage. A bet must often contain value hands, bluffs, protection hands, and blocker-driven candidates in proportions that prevent profitable counter-strategies. Consequently, a hand may bet not because it is locally strong, but because it occupies a necessary role inside the betting range; conversely, a stronger hand may check because its showdown value realizes well and the betting line needs weaker bluff candidates for balance. The decision boundary is therefore determined jointly by private cards, public board texture, both players' ranges, blocker effects, and continuation values. This is exactly the kind of coupled hidden-state reasoning that CFR-style and search-based poker solvers are designed to approximate \citep{zinkevich2007regret,brown2017safe,brown2020combining}.

\paragraph{Solver outputs are precise but not communicable.}
Solver outputs provide the desired equilibrium target, but not in a directly communicable form. A solver table gives action frequencies and continuation values for many private hands under a fixed public state; it does not state the compact contrastive rule explaining why two similar hands diverge. An LLM can therefore produce a plausible one-hand rationale while still missing the frequency allocation that makes the whole range balanced. The core task in this paper is to bridge this gap: first distill solver behavior into a sparse strategic representation, and then articulate local counterfactual distinctions that an independent LLM can transfer to unseen hands. This motivates our use of MDT as a solver-grounded intermediate representation and SCCS as the rule-extraction mechanism.

\section{Methodology}
\label{sec:method}

Our goal is to achieve solver-guided articulation, which is a conditional prediction problem with an explicit intermediate representation. At a decision point $h$, let $\pi^*(\cdot\mid h)$ be the computed mixed action distribution, let $\{Q^*(h,a)\}_{a\in\mathcal{A}(h)}$ be the corresponding action-value summaries, and let $\mathbf{x}$ denote a compact representation of public context and continuation summaries. The solver-side object is
\begin{equation}
\mathcal{O}(h)=\bigl(\pi^*(\cdot\mid h),\{Q^*(h,a)\}_{a\in\mathcal{A}(h)},\mathbf{x}\bigr),
\end{equation}
and the articulation procedure produces a rule $r = A(\mathcal{O}(h),\mathcal{D})$, where $\mathcal{D}$ is a reference collection used to locate matched public contexts and policy-divergent comparisons. This collection can be expanded by querying additional solver states, for example by extending action continuations or adding private-hand assignments. The rule works as an intermediate representation supplied to a downstream predictor.

The rule is constrained to use a small set of quantities from $\mathbf{x}$ and comparisons drawn from matched public contexts in $\mathcal{D}$. This rules out explanations that simply restate the full mixed policy table. It also distinguishes articulation from dominant-action labeling: the target remains the full distribution $\pi^*(\cdot\mid h)$, and the intermediate rule must preserve a local mixed-policy distinction rather than only identify the largest-probability action.

For evaluation, the target hand $h_{\mathrm{test}}$ is excluded from the displayed reference examples, and its solver policy is masked from the downstream predictor. SCCS uses the target on the solver side to identify its MDT route and select policy-separated shadow hands, but never displays the target policy. A predictor receives the public scenario, the target hand, and optionally the rule $r$, then outputs a distribution $\tilde{\pi}(\cdot\mid h_{\mathrm{test}})$. The primary metric is the distance between $\tilde{\pi}$ and the masked solver target $\pi^*(\cdot\mid h_{\mathrm{test}})$, compared against direct prompting and prompting with raw summaries alone. Under this formulation, a rule is useful only if it improves distributional prediction on an unseen target rather than copying a displayed policy.

\subsection{Mixed-Strategy Decision Tree}
\label{sec:MDT_arch}

The main technique is a sparse mixed-strategy distillation model, which we call the Mixed-Strategy Decision Tree (MDT; Figure~\ref{fig:pipeline}). For each solver-labeled decision point, we write the MDT input as $\mathbf{x}$. It contains public context (board, action history, position, stacks, and pot information) together with line-conditioned range-level and hand-level continuation summaries derived from solver outputs. These summaries include EV and EQ quantities and action-gap quantities under available lines. In this way, $\mathbf{x}$ captures decision-relevant continuation information in a compact form for solver-policy articulation.

Given strategic-summary input $\mathbf{x}$, each leaf $l\in\mathcal{L}$ stores a pure action $a_l\in\mathcal{A}$. The router induces a probability mass over leaves,
\begin{equation}
\rho_l(\mathbf{x}) =
\prod_{(n,c)\in \mathrm{Path}(l)} p_{n,c}(\mathbf{x}),
\end{equation}
and the distilled mixed policy is obtained by aggregating the mass of leaves assigned to each action:
\begin{equation}
\hat{\pi}(a\mid \mathbf{x}) =
\sum_{l\in\mathcal{L}} \rho_l(\mathbf{x}) \mathbf{1}[a_l=a].
\end{equation}
Thus, MDT represents mixed strategies through probabilistic routing over pure-action leaves, rather than by storing a mixed action distribution inside each leaf. A leaf never carries a full action-frequency vector; it names one action prototype. The leaves remain intentionally simple, while the routing structure encodes when each pure strategic action should receive probability mass. The final model uses hard sparse local routers; implementation details and router ablations are provided in Appendix~\ref{app:router}.

We train the MDT against solver policy labels using a composite objective. Let $\pi^*(\cdot \mid h)$ denote the oracle action distribution and let $Q^*(h,a)$ denote solver-provided action values. We use
$\mathcal{L}_{\mathrm{task}} = \lambda_{\pi}\mathcal{L}_{L_1} + \lambda_{\mathrm{ev}}\mathcal{L}_{\mathrm{EV}}$,
with
\begin{equation}
\mathcal{L}_{L_1} = \frac{1}{|\mathcal{A}|}\sum_{a \in \mathcal{A}} |\pi^*(a \mid h) - \hat{\pi}(a \mid h)|,
\end{equation}
\begin{equation}
\mathcal{L}_{\mathrm{EV}} =
V^*(h) -
\sum_{a \in \mathcal{A}} \hat{\pi}(a \mid h)Q^*(h,a),
\qquad
V^*(h)=
\sum_{a \in \mathcal{A}} \pi^*(a \mid h)Q^*(h,a).
\end{equation}
Here $\mathcal{L}_{L_1}$ measures fidelity to the oracle mixing frequencies, while $\mathcal{L}_{\mathrm{EV}}$ measures the oracle-conditioned EV gap under solver-provided action values. We treat this strictly as a local fidelity measure, not as a full-game exploitability estimate.

There are two optimization strategies.
The first strategy keeps the router differentiable and adds sparsity-inducing regularization:
$\mathcal{L}_{\mathrm{soft}} = \mathcal{L}_{\mathrm{task}} + \lambda_1 \|\mathbf{w}\|_1 + \lambda_2 \mathcal{H}(\mathbf{p}) + \lambda_3 \mathcal{L}_{\mathrm{ortho}}$.
This regime generally achieves better fidelity, but in practice it leaves a long tail of weak summaries with non-zero influence. As a result, the visible top summaries do not fully explain the final decision.
The second strategy progressively converts the router into a strict top-$K$ sparse structure. We use a teacher-assisted curriculum: a dense teacher first absorbs the raw mapping, then a structured student inherits topology and is finally locked into hard summary selection with $K=5$ active local summaries per node. The goal is not maximal fidelity, but explicitness: the displayed sparse summaries are the variables used to compute the router probabilities. We compare both approaches in the experiments.

\subsection{Scenario-Constrained Counterfactual Sampling}
\label{sec:sccs}

The MDT exposes which sparse summaries are used along a route, but a routed path alone is not yet a communicable rule. A path can indicate that a hand depends on draw strength, kicker quality, hand EV, or range-relative equity, but it does not identify which local change would move the hand across a strategic boundary. This distinction matters in poker because strategically nearby hands can share the same public context and similar raw equity while occupying different roles inside the equilibrium range.

SCCS addresses this gap by explaining a target hand contrastively. Instead of describing the hand in isolation, it selects a shadow hand from a reference collection of solver queries in the same public scenario whose solver policy diverges and whose MDT route crosses a critical branching decision. For example, under the same board and betting line, a weak-kicker draw may be used as a semi-bluff, while a higher-showdown-value version may check because it realizes enough equity without building the pot. By fixing the public scenario and varying only the private hand, SCCS isolates the strategic quantity that changes the hand's role.

For a target state $h$, SCCS identifies its trained-tree route, finds policy-divergent shadow hands under the same public context, localizes the routing boundary where the target and shadow diverge, and converts the active-summary contrast into a natural-language rule. The procedure is targeted at policy-divergent samples rather than nearest visual or lexical neighbors; details are given in Algorithm~\ref{alg:sccs} and Appendix~\ref{app:sccs_details}. The resulting rule also defines a communicability test: an independent reasoner should be able to apply the extracted local distinction to an unseen target state with matched public context.

\section{Experiments and Results}
\label{sec:exp_results}

The NLH distillation and communicability evaluation aims to answer three questions. First, can the proposed MDT articulate the solver output to improve LLM reasoning? Second, do the rules extracted by SCCS help independent LLMs use solver-derived distinctions on unseen target hands in matched public contexts? Third, what kind of strategic distinction does the rule expose in an individual case?

\subsection{Solver Oracle Interface and Evaluation Samples}

Our experiments use over 250 million solver-labeled postflop decision samples obtained by querying a commercial, high-end NLH solver oracle, including mixed policies and continuation-value quantities. These samples comprise approximately 16M flop decisions and 235M turn decisions under a 6-player 100BB no-rake NLH configuration. Decision points involve postflop spots where two players are left in the game. We include five preflop configurations: SB vs BB single-raised pot and 3-bet pot, BTN vs BB single-raised pot and 3-bet pot, and BTN vs SB 3-bet pot. The same pipeline can continue to generate additional labeled states for new board textures, action branches, private-hand assignments, and configuration choices.

To ensure coverage of strategically distinct public states, flop data is sampled from 1,755 strategic board textures. For each board we record multiple canonical action nodes, including the root, check line, bet line, bet-call line, and bet-bet line. Turn data is generated by extending representative flop branches such as check-check, bet-call, and check-bet-call, then sampling five turn cards per branch. Bet sizes follow the solver's abstraction; all-in actions are folded into the generic ``bet'' action category.

For communicability, we construct matched-context tests with unseen target hands. Each test fixes the public context and asks an independent LLM to predict the solver-equilibrium mixed strategy for an unseen target hand. We compare four prompting conditions: \textbf{Direct}, which provides only the public state and hand; \textbf{Direct+Summaries}, which additionally provides the numerical summaries; \textbf{Route-only}, which displays the target hand's MDT route trace but omits every reference/shadow hand and cross-hand comparison; and \textbf{SCCS Rule}, which adds a contrastive rule extracted from policy-divergent shadow hands in the trained MDT. Route-only therefore controls for exposing the tree computation without the matched counterfactual contrast. To prevent the SCCS rule from serving as a near-label lookup, the target hand is never included among the rule-displayed reference/shadow hands, and its solver policy remains hidden from the LLM. Its solver policy must also differ from the policy of every rule-displayed hand by at least $0.20$ under the same action-averaged $L_1$ metric. Thus, the rule cannot be applied by copying a displayed strategy from a near-duplicate hand; improvement requires transferring the extracted strategic distinction to the held-out target. We report $L_1$ both to the solver target and to the distilled MDT policy.
Additional construction details, including the SCCS matching criteria and prompt format, are reported in the appendix~\ref{app:prompt}.

\subsection{Training Loss and Distillation Fidelity}

\begin{table}[t]
\caption{\textbf{Training loss and distillation fidelity on the NLH solver-labeled evaluation set.} Lower $L_1$ and oracle-conditioned EV gap are better. The oracle row is zero under this metric by construction. Oracle EV Gap is a local fidelity metric under solver-provided action values, not full-game exploitability. The final hard MDT is the model used for communicable rule extraction.}
\label{tab:training_loss}
\begin{center}
\begin{small}
\begin{sc}
\resizebox{\textwidth}{!}{\begin{tabular}{llcccc}
\toprule
\textbf{Representation} & \textbf{Model} & \textbf{$L_1$ Loss} & \textbf{Oracle EV Gap, \% of pot} & \textbf{Architecture / approx. params} & \textbf{Visibility} \\
\midrule
\textit{-} & \textit{Oracle} & \textit{-} & \textit{0.00\%} & \textit{-} & \textit{N/A} \\
\midrule
\textbf{Raw PBS} & 2-Layer MLP & $0.068 \pm 0.002$ & $0.55 \pm 0.04\%$ & 512 hidden; $\sim$1.02M & Black-box policy \\
& 8-Layer ResNet & $0.036 \pm 0.001$ & $0.18 \pm 0.02\%$ & 512 hidden; $\sim$2.60M & Black-box policy \\
\midrule
\textbf{Strategic summaries} & 1-Layer MLP & $0.098 \pm 0.010$ & $0.72 \pm 0.05\%$ & 256 hidden; $\sim$41K & Dense summary model \\
& 2-Layer MLP & $0.053 \pm 0.005$ & $0.37 \pm 0.03\%$ & 256 hidden; $\sim$107K & Dense summary model \\
& 4-Layer MLP & $0.045 \pm 0.005$ & $0.26 \pm 0.03\%$ & 256 hidden; $\sim$238K & Dense summary model \\
& 8-Layer ResNet & $0.021 \pm 0.003$ & $0.05 \pm 0.01\%$ & 256 hidden; $\sim$502K & Dense summary model \\
\midrule
\textbf{Tree students} & Dense-router tree & $0.031 \pm 0.004$ & $0.12 \pm 0.03\%$ & $\sim$19K & Tree, dense routing \\
(Ablations) & Soft-sparse tree & $0.057 \pm 0.005$ & $0.21 \pm 0.02\%$ & Sparsity reg.; $\sim$19K & Tree, regularized routing \\
\midrule
\textbf{Hard curriculum} & Teacher 1 (global sparse) & $0.028 \pm 0.002$ & $0.07 \pm 0.01\%$ & 8-layer ResNet; $\sim$502K & Global summary gate \\
& Teacher 2 (local sparse) & $0.044 \pm 0.005$ & $0.23 \pm 0.03\%$ & 2-Layer MLP routers; $\sim$1.64M & Local tree routing \\
& Final hard MDT & $0.087 \pm 0.007$ & $0.44 \pm 0.04\%$ & top-k/node mask; $\sim$0.7K\\
\bottomrule
\end{tabular}}
\end{sc}
\end{small}
\end{center}
\end{table}

The distillation results motivate MDT as an articulation layer rather than only a predictor. Within the same strategic-summary input, tree-structured routing fits the solver policy substantially better than flat MLPs at comparable or smaller parameter counts in our experiments, suggesting that mixed-equilibrium decisions benefit from a hierarchical representation: coarse public-state and range-level conditions first select a strategic region, while hand-level summaries define local action-frequency boundaries. The hard sparse MDT sacrifices some fidelity by allowing only a small set of summaries at each node, but this constraint makes the routing boundaries explicit enough for SCCS rule extraction. Full training-loss comparisons and EV-gap analysis are reported in Appendix~\ref{app:distillation_fidelity}.

\subsection{Communicability on Unseen Target Hands}

Table~\ref{tab:comm} is the main quantitative result. Across eight LLM configurations, SCCS rules reduce average $L_1$ to the solver target from $0.211$ to $0.100$, a $52.6\%$ relative improvement over direct prompting. The distance to the distilled MDT policy falls from $0.204$ to $0.114$, a $44.0\%$ relative improvement. The SCCS columns also show low variation across the evaluated LLM configurations, suggesting that the effect is not specific to one model setting. SCCS also improves on Route-only: $L_1$ falls from $0.173$ to $0.100$ relative to the solver and from $0.172$ to $0.114$ relative to MDT, reductions of $42.2\%$ and $33.7\%$, respectively. The SCCS point estimate is lower than Route-only in all eight configurations under both targets. Argmax-action agreement, defined as whether the highest-probability predicted action matches the highest-probability solver action, also rises from $57.2\%$ to $76.1\%$ from Direct to SCCS.

The summaries-only condition is intentionally included as a negative control. Direct+\allowbreak Summaries has worse average $L_1$ to the solver target ($0.256$) than the direct prompt, suggesting that raw continuation quantities are not automatically communicable to an LLM. Without a contrastive rule, the model may not know which summary changes are decision-relevant in the current scenario; the extra quantities can be treated as noise or can reinforce a locally conservative interpretation. SCCS improves performance because it organizes those quantities around a policy-divergent boundary under the same public context.

\begin{table}[t]
\caption{\textbf{Communicability on unseen target hands in matched public contexts.} Lower $L_1$ is better. Each model entry reports mean $\pm$ standard error over unseen target cases. The Mean row reports mean $\pm$ sample standard deviation across eight LLM configurations.}
\label{tab:comm}
\centering
\begin{scriptsize}
\resizebox{\textwidth}{!}{\begin{tabular}{lcccc|cccc}
\toprule
\multirow{2}{*}{\textbf{LLM run}} & \multicolumn{4}{c|}{\textbf{$L_1$ to solver}} & \multicolumn{4}{c}{\textbf{$L_1$ to MDT}}\\
& Direct & +Summaries & +Route-only & +SCCS Rule & Direct & +Summaries & +Route-only & +SCCS Rule\\
\midrule
Gemini-3.1 Flash & $0.211 \pm 0.015$ & $0.223 \pm 0.016$ & $0.144 \pm 0.010$ & $0.099 \pm 0.009$ & $0.204 \pm 0.014$ & $0.224 \pm 0.015$ & $0.142 \pm 0.009$ & $0.114 \pm 0.009$ \\
Gemini-3.1 Pro low & $0.235 \pm 0.020$ & $0.299 \pm 0.021$ & $0.228 \pm 0.015$ & $0.106 \pm 0.010$ & $0.235 \pm 0.020$ & $0.300 \pm 0.020$ & $0.234 \pm 0.014$ & $0.124 \pm 0.010$ \\
Gemini-3.1 Pro high & $0.268 \pm 0.019$ & $0.303 \pm 0.015$ & $0.205 \pm 0.015$ & $0.089 \pm 0.012$ & $0.261 \pm 0.019$ & $0.314 \pm 0.012$ & $0.209 \pm 0.015$ & $0.109 \pm 0.013$ \\
DeepSeek-V4 Flash & $0.185 \pm 0.015$ & $0.263 \pm 0.019$ & $0.212 \pm 0.014$ & $0.116 \pm 0.013$ & $0.173 \pm 0.013$ & $0.256 \pm 0.018$ & $0.207 \pm 0.013$ & $0.130 \pm 0.012$ \\
DeepSeek-V4 Pro & $0.254 \pm 0.016$ & $0.255 \pm 0.017$ & $0.174 \pm 0.011$ & $0.112 \pm 0.011$ & $0.252 \pm 0.014$ & $0.259 \pm 0.014$ & $0.171 \pm 0.010$ & $0.125 \pm 0.010$ \\
GPT-5.4 & $0.187 \pm 0.013$ & $0.216 \pm 0.014$ & $0.124 \pm 0.009$ & $0.095 \pm 0.010$ & $0.175 \pm 0.011$ & $0.210 \pm 0.014$ & $0.119 \pm 0.008$ & $0.104 \pm 0.008$ \\
GPT-5.5 low & $0.158 \pm 0.013$ & $0.241 \pm 0.016$ & $0.150 \pm 0.010$ & $0.096 \pm 0.009$ & $0.155 \pm 0.011$ & $0.241 \pm 0.016$ & $0.149 \pm 0.009$ & $0.109 \pm 0.009$ \\
GPT-5.5 high & $0.186 \pm 0.007$ & $0.248 \pm 0.017$ & $0.147 \pm 0.009$ & $0.086 \pm 0.010$ & $0.173 \pm 0.008$ & $0.260 \pm 0.015$ & $0.147 \pm 0.008$ & $0.097 \pm 0.008$\\
\midrule
Mean & $0.211 \pm 0.038$ & $0.256 \pm 0.032$ & $0.173 \pm 0.038$ & $\mathbf{0.100 \pm 0.011}$ & $0.204 \pm 0.041$ & $0.258 \pm 0.035$ & $0.172 \pm 0.040$ & $\mathbf{0.114 \pm 0.011}$\\
\bottomrule
\end{tabular}}
\end{scriptsize}
\end{table}

The table supports a specific interpretation of the result. SCCS does not make an LLM a standalone poker agent; it makes a local solver distinction communicable enough for an independent model to transfer the extracted rule to an unseen target hand in a similar strategic neighborhood. The Route-only comparison further shows that exposing the target route is useful, but the matched shadow contrast communicates additional information about how probability mass moves across the boundary. In a post-hoc matched diagnostic restricted to predictions where Direct, Route-only, and SCCS all select the correct solver argmax, SCCS still lowers Route-only $L_1$ by $38.8\%$ to the solver and $33.0\%$ to MDT. Thus, the difference is not explained only by correcting the dominant action; it includes the remaining probability allocation. Appendix~\ref{app:route_only} gives the construction, paired diagnostics, and parse-success accounting.

\subsection{Case Study: From Over-Folding to Draw-Aware Mixing}

To qualitatively inspect the strategy acquired by the LLM, we inspect one of the most difficult spots for humans: SB vs BB single raised pot. Table~\ref{tab:case_study} in the appendix reports the case where the public state is a \texttt{8s6h5d} board after the bet--raise actions. Now the out-of-position player (SB) is facing pressure with \texttt{Td9c}.

% Directly prompting an LLM identifies the hand as an offsuit gutshot with poor equity realization and assigns most mass to Fold. Despite that the LLM is able to correctly identify the hand bucket, the solver disagrees with its action and assigns no fold mass and mixes between Call and the Raise action. A studied human player will be able to understand that at a relatively high stack depth, a hand like \texttt{T9} can hit the absolute nuts on a \texttt{8} turn or river. At the same time, with the wide range of SB vs BB, the overcards \texttt{J} and \texttt{T} provides additional equity even without the backdoor flush draw. SCCS makes this distinction communicable by contrasting a folding prototype (\texttt{Jc4c}) with the same gut shot bucket. Their differences in draw strength, hand EV, and range-relative equity move \texttt{Td9c} into a continuing draw class with some aggressive raise frequency.

Directly prompting an LLM identifies the hand as an offsuit gutshot with poor equity realization and assigns most mass to Fold. Despite that the LLM is able to correctly identify the hand bucket, the solver disagrees with its action and assigns no fold mass and mixes between Call and the Raise action. A studied human player understands that at relatively deep stack depths, it is crucial to continue to the turn with \texttt{T9} to ensure that the absolute nuts remain in their range on \texttt{7} turn or river runouts. This holds true even if a naive calculation of immediate pot odds suggests the call is unprofitable. At the same time, with the wide range of SB vs BB, the overcards \texttt{J} and \texttt{T} provides additional equity even without the backdoor flush draw. SCCS makes this distinction communicable by contrasting a folding prototype (\texttt{Jc4c}) with the same gut shot bucket. Their differences in draw strength, hand EV, and range-relative equity move \texttt{Td9c} into a continuing draw class with some aggressive raise frequency.

This example illustrates why conservative heuristics are not enough. The direct and summaries-only prompts over-fold, while the MDT rule exposes the local boundary between weak folding draws and continuing draw candidates. 
% With that boundary exposed, GPT-5.5 high shifts toward the solver's Call/Bet allocation. 
With that boundary exposed, GPT-5.5 high shifts toward the solver's Call/Bet mixture.
Appendix~\ref{app:case_study_details} provides the full case details.

\subsection{Exact Strategic Evaluation in Complete River Endgames}
\label{sec:river_eval}

The local $L_1$ and oracle-conditioned EV gap measure fidelity at sampled decisions. To complement them with a complete-policy evaluation, we construct two full heads-up NLH River endgames initialized from Subgames 3 and 4 released by \citet{brown2019solving}. The two public states begin with root pots of $5$ and $37.5$ big blinds, respectively. From each state, our two-size action tree proceeds through fold or showdown. Soft and Hard MDT use the same depth-4 ternary probabilistic-routing architecture with pure-action leaves; Soft may use all eligible summaries, while Hard locks each router to five summaries. We then compute exact exploitability within each River tree, defined here as the average additional payoff available when each player independently switches to its best response.

\begin{table}[t]
\caption{\textbf{Exact strategic evaluation in two complete local River endgames.} Values are milli-big-blinds per game (mbb/g), followed by exploitability as a percentage of the root pot. Hard--Soft reports the observed difference associated with strict Top-5 sparsity. Deep CFR is a same-tree finite-training reference. Lower is better.}
\label{tab:river_exploitability}
\centering
\begin{small}
\resizebox{\textwidth}{!}{\begin{tabular}{lrrrr}
\toprule
\textbf{River state} & \textbf{Soft MDT} & \textbf{Hard Top-5 MDT} & \textbf{Hard--Soft} & \textbf{Deep CFR} \\
\midrule
Subgame 3 ($5$ bb pot) & $6.691$ ($0.1338\%$) & $17.173$ ($0.3435\%$) & $10.482$ ($0.2096$ pp) & $43.075$ ($0.8615\%$) \\
Subgame 4 ($37.5$ bb pot) & $55.013$ ($0.1467\%$) & $133.046$ ($0.3548\%$) & $78.033$ ($0.2081$ pp) & $133.495$ ($0.3560\%$) \\
\bottomrule
\end{tabular}}
\end{small}
\end{table}

The larger absolute values in Subgame 4 reflect its $7.5\times$ larger root pot. After normalization, the two states agree closely: Soft costs $0.1338$--$0.1467\%$ of the root pot and Hard Top-5 costs $0.3435$--$0.3548\%$, making the measured Hard--Soft difference approximately $0.21$ percentage points in both endgames. For scale, we train Deep CFR \citep{brown2019deepcfr} on the identical trees with $10{,}000$ sampled traversals per player per iteration; Table~\ref{tab:river_exploitability} reports its linearly weighted average strategy at iteration 800. This is a matched finite-compute reference rather than a convergence limit. The MDT policies are trained and strategically selected on these same trees, so this experiment measures in-domain strategic compression rather than cross-game generalization. Appendix~\ref{app:river_eval} provides the states, fixed-target value losses, policy errors, training and selection details, and metric definition.

\subsection{Beyond NLH: Liar's Dice}
\label{sec:liars_dice}

We additionally instantiate the solver--MDT--SCCS pipeline in two-player Liar's Dice, with two three-faced dice per player and bids up to quantity three. Each player has six unordered private dice types. A full-tree CFR+ solver supplies the approximate-Nash target, and six-fold cross-fitting holds out one private type per public context. MDT uses 17 summaries of the public history, posterior belief, private dice, and equity; per-action continuation values and Q-gaps are neither MDT inputs nor shown in the prompts.

\begin{table}[t]
\caption{\textbf{Exploratory Liar's Dice communicability results.} Entries are unweighted means across eight LLM configurations; each configuration mean is computed over its successfully parsed outputs. No missing output is imputed. Lower $L_1$ is better. Full per-model results and parse coverage are in Appendix~\ref{app:liars_dice}.}
\label{tab:liars_dice_comm}
\centering
\begin{small}
\begin{tabular}{lcc}
\toprule
\textbf{Prompt condition} & \textbf{$L_1$ to solver} & \textbf{$L_1$ to MDT} \\
\midrule
Direct & $0.174$ & $0.187$ \\
Direct+Summaries & $0.171$ & $0.174$ \\
Route-only & $0.116$ & $0.082$ \\
SCCS Rule & $\mathbf{0.105}$ & $\mathbf{0.079}$ \\
\bottomrule
\end{tabular}
\end{small}
\end{table}

On this exploratory surface, SCCS lowers solver $L_1$ by $39.7\%$ relative to Direct and has the lowest descriptive mean under both targets. Its incremental improvement over Route-only is smaller than in NLH. One plausible reason is the much sparser within-context comparison space: Liar's Dice has only six private types per public context, while an NLH range contains hundreds of feasible private hands. After imposing the same-context, route-divergence, and policy-divergence constraints, fewer nearby shadows remain, so the target route already conveys more of the available local information. This explanation is consistent with the observed setting but is not a causal game-size ablation.

We also assemble one held-out MDT prediction for every Liar's Dice information set from the six cross-fitting folds and evaluate the resulting policy against the fixed approximate-Nash opponent over the complete tree. The seat-averaged value loss is $0.058$ per game on the $[-1,+1]$ terminal payoff scale. Together with the NLH result, this experiment shows that the computational interface is not tied to poker cards, poker vocabulary, or a commercial poker solver. Appendix~\ref{app:liars_dice} records the exploratory selection, partial-response and parse-success accounting, per-configuration results, and fully reproducible game/solver setup.

\section{Scope and Broader Applications}
\label{sec:broader_scope}

The empirical claim of this paper is deliberately specific: solver-derived rules improve an independent LLM's inference of held-out mixed-Nash policies in imperfect-information games. SCCS is especially natural in this setting because it can hold the public situation fixed while contrasting feasible private states. The compact MDT representation is a human-readable communication layer over a large solver state, rather than an assumption that the underlying task is intrinsically low-dimensional.

The broader solver-to-LLM paradigm requires three conditions. First, a reliable solver, planner, search procedure, or simulator-backed optimizer must produce decisions and evaluations that are trustworthy or independently verifiable. Second, the task must be difficult enough that human demonstrations are incomplete, unreliable, or systematically below the desired solution quality. Third, the solver must generate an abundant stream of solved states, trajectories, alternatives, or evaluations. Under these conditions, solver-generated supervision can expose structure that is absent from human records rather than merely imitate the human-data ceiling.

The closest direct extensions are other solver-backed imperfect-information games. In Bayesian security and cyber-defense games, equilibrium solvers can vary private threat signals or attacker types under the same public constraints; articulated contrasts could strengthen an LLM's reasoning about how hidden-risk beliefs change mixed defense allocations. In discretized auctions or bargaining games, private valuations or reservation values play the role of hidden types; solver-derived rules could teach how those types change equilibrium bid or offer distributions. These settings retain the public/private decomposition used by SCCS, although their domain summaries would differ from poker's.

The paradigm is potentially broader than the present SCCS construction. In chess and Go, superhuman engines can generate effectively unlimited positions, policies, and values; extracted rules could strengthen long-horizon planning and transfer of engine-discovered strategic patterns to unseen positions \citep{silver2017mastering,silver2018general}. In weather forecasting, validated numerical and ensemble systems generate trajectories under controlled initial-condition changes \citep{lam2023graphcast}; their outputs could strengthen LLMs' physical, causal, and probabilistic reasoning, including tracing how interacting pressure, moisture, and wind patterns alter extreme-event risk and expressing calibrated uncertainty. In chip and circuit design, electronic-design-automation optimizers, simulators, and formal verification explore designs that humans cannot enumerate \citep{mirhoseini2021chip}; solver-derived rules could strengthen LLMs' constraint-aware multi-objective reasoning, failure diagnosis, and verifiable design refinement across timing, power, area, and stability. These are prospective applications of the broader paradigm, not claims that SCCS transfers unchanged: other domains may require different intermediate representations and counterfactual constructions.

\section{Conclusion}
This paper studies how to make solver-computed mixed strategies usable by LLMs through communicable rules. Human play, commentary, and human-generated rationales provide weak supervision for complex games because they are selective, heuristic, and biased toward pure actions, whereas equilibrium play in imperfect-information games requires precise mixed-strategy frequency allocation. Solver outputs provide the desired optimality signal, but only as numerical policy distributions and continuation values.

We introduced Mixed-Strategy Decision Tree (MDT) to articulate solver-implied decision logic into sparse strategic rules, and Scenario-Constrained Counterfactual Sampling (SCCS) to expose local contrastive boundaries between hands in the same public context. In NLH, using over 250 million solver-labeled mixed-strategy decisions, the resulting rules reduce LLM action-averaged $L_1$ distance to the solver target by $52.6\%$ across eight LLM configurations. These results suggest that solver-generated data can serve not only as supervision for policy prediction, but also as a source of readable reasoning traces for LLMs. The Route-only ablation, complete River-endgame evaluations, and fully releasable Liar's Dice game/solver pipeline with archived outputs further measure the contribution of the contrastive rule, the strategic cost of strict Top-5 sparsity, and portability to another imperfect-information mixed-equilibrium game.

\bibliographystyle{plainnat}
\bibliography{references}

@article{zinkevich2007regret,
  title={Regret minimization in games with incomplete information},
  author={Zinkevich, Martin and Johanson, Michael and Bowling, Michael and Piccione, Carmelo},
  journal={Advances in neural information processing systems},
  volume={20},
  year={2007}
}

@article{bowling2015heads,
  title={Heads-up limit hold'em poker is solved},
  author={Bowling, Michael and Burch, Neil and Johanson, Michael and Tammelin, Oskari},
  journal={Science},
  volume={347},
  number={6218},
  pages={145--149},
  year={2015}
}

@article{brown2018superhuman,
  title={Superhuman AI for heads-up no-limit poker: Libratus beats top professionals},
  author={Brown, Noam and Sandholm, Tuomas},
  journal={Science},
  volume={359},
  number={6374},
  pages={418--424},
  year={2018}
}

@article{brown2019superhuman,
  title={Superhuman AI for multiplayer poker},
  author={Brown, Noam and Sandholm, Tuomas},
  journal={Science},
  volume={365},
  number={6456},
  pages={885--890},
  year={2019}
}

@article{brown2020combining,
  title={Combining deep reinforcement learning and search for imperfect-information games},
  author={Brown, Noam and Bakhtin, Anton and Lerer, Adam and Gong, Qucheng},
  journal={Advances in neural information processing systems},
  volume={33},
  pages={17057--17069},
  year={2020}
}

@article{gupta2023chatgpt,
  title={Are ChatGPT and GPT-4 good poker players? A pre-flop analysis},
  author={Gupta, Akshat},
  journal={arXiv preprint arXiv:2308.12466},
  year={2023}
}

@article{huang2024pokergpt,
  title={PokerGPT: An end-to-end lightweight solver for multi-player Texas Hold'em via large language model},
  author={Huang, Chenghao and Cao, Yanbo and Wen, Yinlong and Zhou, Tao and Zhang, Yanru},
  journal={arXiv preprint arXiv:2401.06781},
  year={2024}
}

@article{hinton2015distilling,
  title={Distilling the knowledge in a neural network},
  author={Hinton, Geoffrey and Vinyals, Oriol and Dean, Jeff},
  journal={arXiv preprint arXiv:1503.02531},
  year={2015}
}

@article{frosst2017distilling,
  title={Distilling a neural network into a soft decision tree},
  author={Frosst, Nicholas and Hinton, Geoffrey},
  journal={arXiv preprint arXiv:1711.09784},
  year={2017}
}

@article{agarwal2021neural,
  title={Neural additive models: Interpretable machine learning with neural nets},
  author={Agarwal, Rishabh and Melnick, Levi and Frosst, Nicholas and Zhang, Xuezhou and Lengerich, Ben and Caruana, Rich and Hinton, Geoffrey E},
  journal={Advances in Neural Information Processing Systems},
  volume={34},
  pages={4699--4711},
  year={2021}
}

@article{bastani2018verifiable,
  title={Verifiable reinforcement learning via policy extraction},
  author={Bastani, Osbert and Pu, Yewen and Solar-Lezama, Armando},
  journal={Advances in neural information processing systems},
  volume={31},
  year={2018}
}

@inproceedings{verma2018programmatically,
  title={Programmatically interpretable reinforcement learning},
  author={Verma, Abhinav and Murali, Vijayaraghavan and Singh, Rishabh and Kohli, Pushmeet and Chaudhuri, Swarat},
  booktitle={International Conference on Machine Learning},
  pages={5045--5054},
  year={2018}
}

@inproceedings{koh2020concept,
  title={Concept bottleneck models},
  author={Koh, Pang Wei and Nguyen, Thao and Tang, Yew Siang and Mussmann, Stephen and Pierson, Emma and Kim, Been and Liang, Percy},
  booktitle={International Conference on Machine Learning},
  pages={5338--5348},
  year={2020}
}

@article{mcgrath2022acquisition,
  title={Acquisition of chess knowledge in AlphaZero},
  author={McGrath, Thomas and Kapishnikov, Andrei and Toma{\v{s}}ev, Nenad and Pearce, Adam and Wattenberg, Martin and Hassabis, Demis and Kim, Been and Paquet, Ulrich and Kramnik, Vladimir},
  journal={Proceedings of the National Academy of Sciences},
  volume={119},
  number={47},
  pages={e2206625119},
  year={2022}
}

@article{moravvcik2017deepstack,
  title={Deepstack: Expert-level artificial intelligence in heads-up no-limit poker},
  author={Morav{\v{c}}{\'\i}k, Matej and Schmid, Martin and Burch, Neil and Lis{\`y}, Viliam and Morrill, Dustin and Bard, Nolan and Davis, Trevor and Waugh, Kevin and Johanson, Michael and Bowling, Michael},
  journal={Science},
  volume={356},
  number={6337},
  pages={508--513},
  year={2017},
  publisher={American Association for the Advancement of Science}
}

@article{silver2025welcome,
  title={Welcome to the era of experience},
  author={Silver, David and Sutton, Richard S},
  journal={Google AI},
  volume={1},
  pages={11},
  year={2025}
}

@article{wachter2017counterfactual,
  title={Counterfactual explanations without opening the black box: Automated decisions and the GDPR},
  author={Wachter, Sandra and Mittelstadt, Brent and Russell, Chris},
  journal={Harv. JL \& Tech.},
  volume={31},
  pages={841},
  year={2017},
  publisher={HeinOnline}
}

@article{silver2017mastering,
  title={Mastering the game of go without human knowledge},
  author={Silver, David and Schrittwieser, Julian and Simonyan, Karen and Antonoglou, Ioannis and Huang, Aja and Guez, Arthur and Hubert, Thomas and Baker, Lucas and Lai, Matthew and Bolton, Adrian and others},
  journal={nature},
  volume={550},
  number={7676},
  pages={354--359},
  year={2017},
  publisher={Nature Publishing Group UK London}
}

@inproceedings{ganzfried2014potential,
  title={Potential-aware imperfect-recall abstraction with earth mover's distance in imperfect-information games},
  author={Ganzfried, Sam and Sandholm, Tuomas},
  booktitle={Proceedings of the AAAI Conference on Artificial Intelligence},
  volume={28},
  year={2014}
}

@inproceedings{li2025efficient,
  title={Efficient online pruning and abstraction for imperfect information extensive-form games},
  author={Li, Boning and Huang, Longbo},
  booktitle={The Thirteenth International Conference on Learning Representations},
  year={2025}
}

@article{rusu2015policy,
  title={Policy distillation},
  author={Rusu, Andrei A and Colmenarejo, Sergio Gomez and Gulcehre, Caglar and Desjardins, Guillaume and Kirkpatrick, James and Pascanu, Razvan and Mnih, Volodymyr and Kavukcuoglu, Koray and Hadsell, Raia},
  journal={arXiv preprint arXiv:1511.06295},
  year={2015}
}

@article{wei2022chain,
  title={Chain-of-thought prompting elicits reasoning in large language models},
  author={Wei, Jason and Wang, Xuezhi and Schuurmans, Dale and Bosma, Maarten and Xia, Fei and Chi, Ed and Le, Quoc V and Zhou, Denny and others},
  journal={Advances in neural information processing systems},
  volume={35},
  pages={24824--24837},
  year={2022}
}

@article{gunasekar2023textbooks,
  title={Textbooks are all you need},
  author={Gunasekar, Suriya and Zhang, Yi and Aneja, Jyoti and Mendes, Caio C{\'e}sar Teodoro and Del Giorno, Allie and Gopi, Sivakanth and Javaheripi, Mojan and Kauffmann, Piero and de Rosa, Gustavo and Saarikivi, Olli and others},
  journal={arXiv preprint arXiv:2306.11644},
  year={2023}
}

@article{silver2016mastering,
  title={Mastering the game of Go with deep neural networks and tree search},
  author={Silver, David and Huang, Aja and Maddison, Chris J and Guez, Arthur and Sifre, Laurent and Van Den Driessche, George and Schrittwieser, Julian and Antonoglou, Ioannis and Panneershelvam, Veda and Lanctot, Marc and others},
  journal={nature},
  volume={529},
  number={7587},
  pages={484--489},
  year={2016},
  publisher={Nature Publishing Group}
}

@article{jacobs1991adaptive,
  title={Adaptive mixtures of local experts},
  author={Jacobs, Robert A and Jordan, Michael I and Nowlan, Steven J and Hinton, Geoffrey E},
  journal={Neural computation},
  volume={3},
  number={1},
  pages={79--87},
  year={1991},
  publisher={MIT Press}
}

@article{shazeer2017outrageously,
  title={Outrageously large neural networks: The sparsely-gated mixture-of-experts layer},
  author={Shazeer, Noam and Mirhoseini, Azalia and Maziarz, Krzysztof and Davis, Andy and Le, Quoc and Hinton, Geoffrey and Dean, Jeff},
  journal={arXiv preprint arXiv:1701.06538},
  year={2017}
}

@article{zelikman2022star,
  title={Star: Bootstrapping reasoning with reasoning},
  author={Zelikman, Eric and Wu, Yuhuai and Mu, Jesse and Goodman, Noah},
  journal={Advances in Neural Information Processing Systems},
  volume={35},
  pages={15476--15488},
  year={2022}
}

@article{nash1951noncooperative,
  title={Non-cooperative games},
  author={Nash, John},
  journal={Annals of Mathematics},
  volume={54},
  number={2},
  pages={286--295},
  year={1951}
}

@book{osborne1994course,
  title={A Course in Game Theory},
  author={Osborne, Martin J. and Rubinstein, Ariel},
  publisher={MIT Press},
  year={1994}
}

@incollection{kuhn1953extensive,
  title={Extensive Games and the Problem of Information},
  author={Kuhn, Harold W.},
  booktitle={Contributions to the Theory of Games II},
  editor={Kuhn, Harold W. and Tucker, Albert W.},
  series={Annals of Mathematics Studies},
  volume={28},
  pages={193--216},
  publisher={Princeton University Press},
  year={1953}
}

@article{brown2017safe,
  title={Safe and nested subgame solving for imperfect-information games},
  author={Brown, Noam and Sandholm, Tuomas},
  journal={Advances in neural information processing systems},
  volume={30},
  year={2017}
}

@misc{gtowizardbenchmark2026,
      title={GTO Wizard Benchmark},
      author={Marc-Antoine Provost and Nejc Ilenic and Christopher Solinas and Philippe Beardsell},
      year={2026},
      eprint={2603.23660},
      archivePrefix={arXiv},
      primaryClass={cs.AI},
      url={https://arxiv.org/abs/2603.23660},
}

@inproceedings{zhuang2025pokerbench,
  title={Pokerbench: Training large language models to become professional poker players},
  author={Zhuang, Richard and Gupta, Akshat and Yang, Richard and Rahane, Aniket and Li, Zhengyu and Anumanchipalli, Gopala},
  booktitle={Proceedings of the AAAI Conference on Artificial Intelligence},
  volume={39},
  pages={26175--26182},
  year={2025}
}

@article{achiam2023gpt,
  title={GPT-4 Technical Report},
  author={Achiam, Josh and Adler, Steven and Agarwal, Sandhini and Ahmad, Lama and Akkaya, Ilge and Aleman, Florencia Leoni and Almeida, Diogo and Altenschmidt, Janko and Altman, Sam and Anadkat, Shyamal and others},
  journal={arXiv preprint arXiv:2303.08774},
  year={2023}
}

@article{silver2018general,
  title={A general reinforcement learning algorithm that masters chess, shogi, and Go through self-play},
  author={Silver, David and Hubert, Thomas and Schrittwieser, Julian and Antonoglou, Ioannis and Lai, Matthew and Guez, Arthur and Lanctot, Marc and Sifre, Laurent and Kumaran, Dharshan and Graepel, Thore and others},
  journal={Science},
  volume={362},
  number={6419},
  pages={1140--1144},
  year={2018},
  doi={10.1126/science.aar6404}
}

@inproceedings{lin2026how,
  title={How Far Are LLMs from Professional Poker Players? Revisiting Game-Theoretic Reasoning with Agentic Tool Use},
  author={Lin, Minhua and Dai, Enyan and Liu, Hui and Tang, Xianfeng and Yan, Yuliang and Dai, Zhenwei and Zeng, Jingying and Zhang, Zhiwei and Wang, Fali and Gao, Hongcheng and Luo, Chen and Zhang, Xiang and He, Qi and Wang, Suhang},
  booktitle={The Fourteenth International Conference on Learning Representations},
  year={2026},
  url={https://openreview.net/forum?id=vV54ShHvGi}
}

@inproceedings{brown2019solving,
  title={Solving Imperfect-Information Games via Discounted Regret Minimization},
  author={Brown, Noam and Sandholm, Tuomas},
  booktitle={Proceedings of the AAAI Conference on Artificial Intelligence},
  volume={33},
  pages={1829--1836},
  year={2019},
  doi={10.1609/aaai.v33i01.33011829}
}

@inproceedings{brown2019deepcfr,
  title={Deep Counterfactual Regret Minimization},
  author={Brown, Noam and Lerer, Adam and Gross, Sam and Sandholm, Tuomas},
  booktitle={Proceedings of the 36th International Conference on Machine Learning},
  series={Proceedings of Machine Learning Research},
  volume={97},
  pages={793--802},
  year={2019},
  publisher={PMLR},
  url={https://proceedings.mlr.press/v97/brown19b.html}
}

@inproceedings{guo2024suspicion,
  title={Suspicion-Agent: Playing Imperfect Information Games with Theory of Mind Aware GPT-4},
  author={Guo, Jiaxian and Yang, Bo and Yoo, Paul and Lin, Bill Yuchen and Iwasawa, Yusuke and Matsuo, Yutaka},
  booktitle={Conference on Language Modeling},
  year={2024},
  url={https://openreview.net/forum?id=F2yGbwXJAi}
}

@article{lam2023graphcast,
  title={Learning skillful medium-range global weather forecasting},
  author={Lam, Remi and Sanchez-Gonzalez, Alvaro and Willson, Matthew and Wirnsberger, Peter and Fortunato, Meire and Alet, Ferran and Ravuri, Suman and Ewalds, Timo and Eaton-Rosen, Zach and Hu, Weihua and others},
  journal={Science},
  volume={382},
  number={6677},
  pages={1416--1421},
  year={2023},
  doi={10.1126/science.adi2336}
}

@article{mirhoseini2021chip,
  title={A graph placement methodology for fast chip design},
  author={Mirhoseini, Azalia and Goldie, Anna and Yazgan, Mustafa and Jiang, Joe Wenjie and Songhori, Ebrahim and Wang, Shen and Lee, Young-Joon and Johnson, Eric and Pathak, Omkar and Nova, Azade and others},
  journal={Nature},
  volume={594},
  pages={207--212},
  year={2021},
  doi={10.1038/s41586-021-03544-w}
}

\clearpage
\appendix

\begin{figure}
    \centering
    \includegraphics[width=1\linewidth]{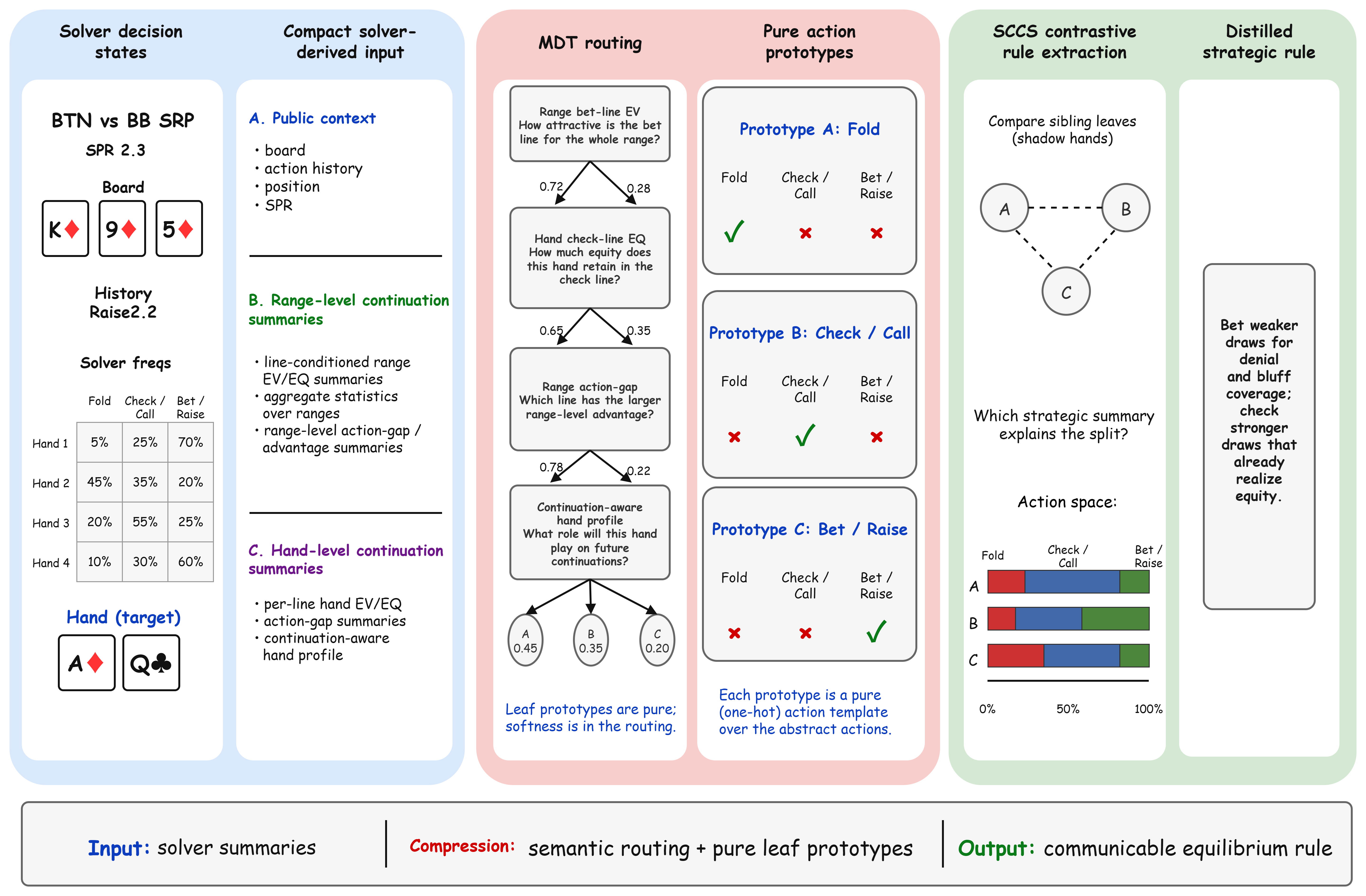}
    \caption{\textbf{Distilling solver policies into communicable strategic rules.} Public context and continuation outputs form the compact MDT input $\mathbf{x}$, while the solver policy supplies the supervised target. MDT uses sparse routing over the input summaries to assign probability mass to pure-action leaf prototypes; the mixed strategy arises from the routing distribution, not from mixed leaves. SCCS compares shadow hands with matched public context but clear solver-policy divergence, then converts the routing contrast into a rule that an independent reasoner can apply to unseen target hands.}
    \label{fig:pipeline}
\end{figure}

\section{Additional Related Work}
\label{app:related_work}

\paragraph{Superhuman solvers as silent oracles.}
The resolution of imperfect-information games has been driven by equilibrium-finding algorithms like Counterfactual Regret Minimization (CFR) and its variants \citep{zinkevich2007regret, bowling2015heads, brown2018superhuman, brown2019superhuman, moravvcik2017deepstack, brown2020combining}. In other strategic domains, superhuman systems have become more than competitors: AlphaGo and AlphaZero changed how strong players and researchers study Go and chess \citep{silver2016mastering,silver2017mastering}, and later analyses recovered human-understandable chess concepts from AlphaZero play \citep{mcgrath2022acquisition}. Poker has undergone a parallel shift toward solver- and GTO-guided study, as reflected by solver-based benchmarks and training resources \citep{zhuang2025pokerbench,gtowizardbenchmark2026}. Yet poker solvers remain largely silent oracles: they provide exact frequency prescriptions (e.g., ``bet 33.4\%'') but not communicable rationales for those frequencies.

\paragraph{Concept discovery and interpretable policy distillation.}
A growing body of work aims to expose structure inside learned or optimized policies. Concept-based analysis has been effective for perfect-information agents such as AlphaZero \citep{mcgrath2022acquisition}, while policy distillation compresses cumbersome teacher models into lightweight students \citep{hinton2015distilling,rusu2015policy}. Interpretable variants \citep{koh2020concept, jacobs1991adaptive, shazeer2017outrageously}, including Tree Imitation Learning, VIPER, and Programmatically Interpretable RL, project policies into trees or programs \citep{bastani2018verifiable,verma2018programmatically}; soft decision trees provide another differentiable route to tree-structured explanations \citep{frosst2017distilling}. In imperfect-information games, abstraction and bucketing methods cluster strategically similar hands for computational tractability \citep{ganzfried2014potential,li2025efficient}. These methods expose useful structure, but their objectives are usually concept probing, compression, or efficient solving; they do not directly address the range-level coupling and action-frequency mixing that make poker equilibrium policies hard to communicate.

\paragraph{LLMs, poker benchmarks, and synthetic reasoning.}
LLMs have recently been tested as poker decision makers, but existing results show substantial gaps. \citet{gupta2023chatgpt} evaluate ChatGPT and GPT-4 on preflop poker decisions, while \citet{huang2024pokergpt} explore LLM-based poker agents trained from online poker data; both lines highlight the difficulty of obtaining reliable strategic behavior from language models alone. PokerBench provides a broader benchmark and training set over curated preflop and postflop spots, using dominant-action and bet-size labels with action-accuracy and exact-match metrics \citep{zhuang2025pokerbench}. The GTO Wizard Benchmark evaluates frontier LLMs against a superhuman poker agent and finds persistent failures in card representation, range construction, and solver-level action selection \citep{gtowizardbenchmark2026}. These works primarily evaluate or train LLM poker play. By contrast, structured reasoning and synthetic-data work suggests that explicit rationales can improve model behavior \citep{wei2022chain,gunasekar2023textbooks,zelikman2022star}, but poker requires such rationales to be grounded in solver-computed mixed strategies rather than in fluent human commentary alone.

\paragraph{Relation to action- and agent-level evaluations.}
PokerGPT reports action prediction, aggregate behavioral statistics, and play against Slumbot; Suspicion-Agent reports finite-match chip outcomes against named agents in Leduc Hold'em \citep{huang2024pokergpt,guo2024suspicion}; and PokerBench scores one dominant action category or an exact bet size and compares agents in head-to-head play \citep{zhuang2025pokerbench}. These are useful evaluations of realized actions and opponent-specific performance, but they do not compare every probability in a solver mixture at matched information sets. The recent GTO Wizard Benchmark further reports, under its separate HUNL protocol, that its approximate-Nash agent beats Slumbot by $194\pm41$ mbb/hand and that GPT-4 loses $1{,}362\pm256$ mbb/hand to that agent \citep{gtowizardbenchmark2026}. Because the games and protocols differ, these numbers are not directly comparable with our $L_1$ results; they illustrate why success against a particular imperfect opponent and fidelity to a Nash mixture are distinct questions. Likewise, action accuracy, exact match, and our argmax agreement all discard the frequencies among supported actions, whereas action-averaged $L_1$ evaluates the complete local distribution.

\section{Distillation Fidelity Details}
\label{app:distillation_fidelity}

Table~\ref{tab:training_loss} reports the training objective components used to evaluate solver-policy distillation: action-distribution $L_1$ and oracle-conditioned EV gap. Dense high-capacity models obtain the lowest numerical loss, but their decisions are not directly inspectable. The raw-PBS baselines use a 1482-dimensional input and 512-wide hidden layers, whereas the summary-input baselines use a 156-dimensional input and 256-wide hidden layers. In the tree variants, the soft-sparse model keeps dense routers over the 156 summaries and relies on regularization rather than hard feature selection; Teacher 2 uses higher-capacity two-layer MLP routers; and the final hard MDT selects its top five summaries at each node through a mask over the original 156-dimensional input. Within the summary-input comparison, tree-structured routing improves substantially over the flat MLP, supporting the use of MDT for context-dependent policy structure. The final hard MDT has higher loss because hard sparsification removes small corrective effects used by dense routers. This fidelity cost is intentional: the displayed node-local summaries are exactly the variables used by the model, which avoids explanations that omit many low-weight contributors.

\section{Case Study Details}
\label{app:case_study_details}

This appendix expands the draw-aware transfer case study from the main text using the GPT-5.5 high visible-rationale diagnostic run. The example is chosen because all methods receive the same public state and target hand, and the main error is strategic rather than notational: \texttt{Td9c} is recognized as an offsuit gutshot, but the question is whether it belongs to the folding part of the range or to a continuing draw class. We report action distributions in the order shown in the public context, and report the average $L_1$ distance to the solver target.

\begin{table}[ht]
% \caption{\textbf{Draw-aware rule transfer on an unseen out-of-position target hand.} The SCCS rule exposes the boundary between weak folding draws and continuing semi-bluff candidates.}
\caption{\textbf{Draw-aware rule transfer on an unseen out-of-position target hand.} The SCCS rule exposes the boundary between weak folding draws and stronger continuing draw candidates.}
\label{tab:case_study}
\centering
\begin{small}
\begin{tabular}{lp{0.68\linewidth}}
\toprule
Public context & SB vs BB single-raised pot, board \texttt{8s6h5d}, history \texttt{b4,b11}, OOP decision, actions \{Call, Fold, Bet($1.79\times$)\}. \\
Contrastive trace & Fold prototype \texttt{Jc4c}: model $[0.02,0.97,0.00]$, solver $[0.00,1.00,0.00]$. Policy-divergent continuing prototypes, including \texttt{Qh9h} and \texttt{Jc7c}, route to Call/Bet mixtures with solver Fold $=0$. The contrastive split is governed by draw strength, hand EV, and range-relative equity. \\
Unseen target hand & \texttt{Td9c}. Key SCCS values: draw strength $0.30$ vs average $0.13$; hand EV $0.03$; hand equity $0.32$; range-relative equity $-0.14$. \\
Solver target & $[\text{Call}:0.63,\ \text{Fold}:0.00,\ \text{Bet}:0.37]$. \\
GPT-5.5 direct & $[0.12,0.88,0.00]$, average $L_1$ to solver $=0.585$. \\
GPT-5.5 + summaries & $[0.07,0.91,0.02]$, average $L_1$ to solver $=0.605$. \\
GPT-5.5 + SCCS rule & $[0.59,0.00,0.41]$, average $L_1$ to solver $=0.025$. \\
\bottomrule
\end{tabular}
\end{small}
\end{table}

% The visible rationales clarify the failure mode. Direct prompting describes \texttt{Td9c} as a gutshot with overcards but poor out-of-position realization, and therefore assigns most mass to Fold. Adding summaries does not change the qualitative decision: the model notes the 9 blocker, but still treats the hand as low-ranked air facing pressure and views the large aggressive action as too ambitious. SCCS changes the evidence structure by presenting a matched folding prototype together with continuing prototypes in the same public context. With this contrast, the model instead anchors \texttt{Td9c} to the nearby T9 gutshot profile: low showdown value but enough straight equity and playability to continue, with suit differences secondary. It therefore removes the fold mass and recovers the solver's Call/Bet mixture up to small error. This is the type of local range-boundary transfer measured in the communicability experiment; it is not an evaluation of live exploitability.

The visible rationales clarify the failure mode. Direct prompting describes \texttt{Td9c} as a gutshot with two overcards but poor out-of-position equity realization, and therefore assigns most of its probability mass to Fold. Adding summaries does not change the qualitative decision: the model notes some relevant card-removal effects, but still treats the hand as low-ranked air facing pressure and views the large aggressive action as too ambitious. SCCS changes the evidence structure by presenting a matched folding prototype together with continuing prototypes in the same public context. This contrast highlights why \texttt{Td9c} belongs to a stronger continuing-draw class: a \texttt{7} on the turn completes the nut straight, a \texttt{J} produces an open-ended straight draw, and a \texttt{T} or \texttt{9} adds pair equity. Despite its low current showdown value, the hand therefore has sufficient draw quality, future-street potential, and range-relative value to continue, with suit differences playing a secondary role. The SCCS-guided model consequently removes the fold mass and recovers the solver's Call/Bet mixture up to a small error. This behavior exemplifies the local range-boundary transfer measured in the communicability experiment; it should not be interpreted as an evaluation of live exploitability.

\section{NLH Rules and Poker Terminology}
\label{app:NLH_terms}

This appendix provides a compact reference for the NLH rule structure and poker terminology used throughout the paper. The descriptions are intended to fix notation and vocabulary for the experiments, not to introduce new modeling assumptions.

\subsection{Rules of  No-Limit Texas Hold'em}

\paragraph{Game format.}
No-Limit Texas Hold'em (NLH) is a form of Texas Hold'em in which a player may bet up to the full remaining stack. Our experiments use a six-player configuration, while each evaluated postflop decision has two players remaining in the hand. Each player receives two private cards, usually called \emph{hole cards}. Up to five public \emph{community cards} are then revealed on the board. At showdown, each remaining player forms the best five-card poker hand using any combination of their two private cards and the five community cards.

\paragraph{Blinds and positions.}
Each hand begins with forced bets called the small blind (SB) and big blind (BB). The button marks the dealer position and moves clockwise between hands; in a six-player game, the two players immediately to its left post the SB and BB. Before the flop, action begins with the active player to the left of the BB. On each postflop street, action begins with the first active player to the left of the button. In the two-player postflop situations studied here, the player acting first is out of position (OOP), while the player acting second is in position (IP).

\paragraph{Betting streets.}
A hand proceeds through four betting rounds, also called \emph{streets}. The \emph{preflop} round occurs after private cards are dealt and before any community card appears. The \emph{flop} reveals three community cards, the \emph{turn} reveals a fourth community card, and the \emph{river} reveals the fifth and final community card. In this paper, the dataset and evaluations focus on postflop decisions, especially flop and turn states.

\paragraph{Legal actions.}
At a decision point, the legal actions depend on the previous betting sequence. A player may \emph{check} if no bet is currently faced, \emph{bet} to put chips into the pot, \emph{call} to match an opponent's bet, \emph{fold} to surrender the pot, or \emph{raise} to increase an existing bet. In no-limit poker, a bet or raise can be any legal size up to the player's remaining stack; an all-in action commits the full remaining stack. Our action abstraction groups available aggressive actions under the generic bet/raise category when reporting solver mixtures.

\paragraph{Pots, stacks, and bet sizes.}
The \emph{pot} is the number of chips currently contested. A player's \emph{stack} is their remaining chips. Stack-to-pot ratio (SPR) is the remaining effective stack divided by the pot and measures how much future betting leverage remains. Bet sizes are often written as fractions or multiples of the pot, e.g., a $0.5\times$ pot bet or a $1.79\times$ pot raise. The experiment configuration uses 100BB starting stacks and no rake.

\paragraph{Hand ranking.}
Texas Hold'em uses the standard poker hand order: high card, one pair, two pair, three of a kind, straight, flush, full house, four of a kind, and straight flush. A \emph{kicker} is a side card used to break ties between otherwise similar made hands, such as top pair with an ace kicker versus top pair with a weaker kicker.

\subsection{Common Poker Terms Used in the Paper}

\begin{longtable}{p{0.22\linewidth}p{0.70\linewidth}}
\toprule
\textbf{Term} & \textbf{Meaning in this paper} \\
\midrule
\endfirsthead
\toprule
\textbf{Term} & \textbf{Meaning in this paper} \\
\midrule
\endhead
\bottomrule
\endfoot
\bottomrule
\endlastfoot

Action line / history & The sequence of previous betting actions and public card events leading to the current decision point. \\
All-in & A bet or raise that commits a player's entire remaining stack. \\
Air & A hand with little or no current showdown value and limited immediate equity. \\
Backdoor draw & A draw that needs favorable cards on both later streets to complete, such as needing both turn and river to make a flush. \\
Bet size & The amount placed into the pot, often normalized by the current pot size. \\
Blocker & A card in a player's hand that removes combinations from the opponent's possible range, often reducing the chance that the opponent holds strong continuing hands. \\
Board & The public community cards visible to both players. \\
Board texture & Strategic properties of the board, such as pairedness, connectedness, straight potential, and flush potential. \\
Bluff & An aggressive action with a hand that is unlikely to be best if called, used to make better hands fold. \\
Call & Matching the current bet to continue in the hand. \\
Check & Passing the action when no bet is faced. \\
Continuation bet & A postflop bet made by the player who was the previous aggressor, commonly abbreviated as c-bet. \\
Draw & A hand that is not currently strong but can improve to a strong hand on later community cards. \\
Equity (EQ) & The probability, or solver-computed share, that a hand or range wins at showdown under the relevant future-card distribution. \\
Expected value (EV) & The expected payoff of a hand, range, or action under the solver's continuation strategy. \\
Fold equity & The value gained from the probability that an opponent folds to an aggressive action. \\
Flush draw & A draw to five cards of the same suit. \\
Gutshot & An inside straight draw that can complete with one specific rank. \\
GTO & Game-Theoretic Optimal; in this paper, an approximate Nash-equilibrium poker strategy computed by a solver. \\
Hand & Usually the player's two private cards, and sometimes the resulting best five-card category depending on context. \\
In position (IP) & The player who acts second on postflop streets. \\
MDF & Minimum defense frequency, a pot-odds-derived threshold describing how often a range must continue to avoid being immediately exploitable by a bet. \\
Mixed strategy & A probability distribution over legal actions at a decision point. \\
Nuts & The strongest possible hand, or class of strongest hands, for the current board. \\
Nut advantage & A range-level advantage in the frequency or equity of nut-class hands. \\
Offsuit & A two-card private hand whose cards have different suits. \\
Open-ended straight draw (OESD) & A straight draw that can complete with a card on either end of the sequence. \\
Out of position (OOP) & The player who acts first on postflop streets. \\
Overcard & A private card higher than every card on the board. \\
Postflop & Any decision after the flop has been dealt; includes flop, turn, and river. \\
Pot odds & The price offered by the pot for calling a bet, usually expressed as a required equity threshold. \\
Protection bet & A bet with a vulnerable made hand or semi-made hand intended to deny equity to hands that can improve. \\
Range & The probability distribution over private hands a player can hold after conditioning on public cards and betting history. \\
Range advantage & A range-level equity or EV edge for one player over the other in the current public state. \\
River & The fifth community card and final betting street. \\
Semi-bluff & A bluffing bet or raise with a hand that can improve to a strong hand on later streets. \\
Set & Three of a kind made with a pocket pair and one matching board card. \\
Showdown value & The ability of a hand to win if betting stops and the hand reaches showdown. \\
Single-raised pot & A pot where the preflop action contains one raise and no 3-bet. \\
Suited & A two-card private hand whose cards share the same suit. \\
Thin value & A value bet with a hand that is ahead of some calling hands but not strong enough to be clearly dominant. \\
Three-bet pot / 3-bet pot & A pot where the preflop action contains a raise and then a re-raise. \\
Trap & A passive action with a very strong hand, used to keep weaker hands or bluffs in the opponent's range. \\
Turn & The fourth community card and the betting street after it is dealt. \\
Unblocker & A card property that leaves the opponent's folding range relatively intact, which can improve bluff quality in some contexts. \\
Value bet & A bet made with a hand expected to be called by worse hands often enough to profit. \\
\end{longtable}

\section{Pilot Study on Public Poker Discourse}
\label{app:forum_sft_pilot}

This appendix gives additional context for the pilot study mentioned in Section~\ref{sec:why-hard}. Two Plus Two (2+2) refers to the Two Plus Two poker forum\footnote{\url{https://forumserver.twoplustwo.com/}}, a long-running public discussion forum for poker strategy, theory, and community discussion. Such forum text is useful for exposing a language model to poker vocabulary, common strategic concepts, and the informal reasoning style used by human players.

In our pilot experiment, we scraped public 2+2 forum discussions and used them to perform supervised fine-tuning (SFT) on an open Qwen model with the standard next-token prediction objective. The goal was not to build the final system in this paper, but to test whether imitation of naturally occurring poker discourse could by itself teach equilibrium-relevant poker reasoning. The resulting model learned to use many terms that appeared frequently in the forum text, such as range, blocker, equity, pot odds, bluff, value bet, and GTO. It also produced more fluent local hand explanations than the base model.

However, the improvement was mostly linguistic and heuristic. In qualitative evaluations, the fine-tuned model still showed shallow understanding of many poker-theoretic concepts that appeared in the training text. It could often repeat the vocabulary of range advantage, blockers, or mixed strategy, but it did not reliably apply these concepts to make decisions close to solver-computed GTO strategies. In particular, it frequently reduced mixed-equilibrium decisions to single-hand narratives, over-relied on visible hand strength, and failed to reason about how one private hand's action frequency is constrained by the rest of the range.

These observations support the distinction made in the main text. Public poker discourse is valuable as a source of terminology and human-readable explanation style, but it does not provide ground-truth equilibrium targets. Therefore, our main pipeline uses solver-derived mixed policies and continuation summaries as the strategic target, while using language only as the medium for articulation.

\section{Additional Method Details}
\label{app:detail}

\subsection{Router Parameterization}
\label{app:router}

Each internal node $n$ computes a branch distribution from a small set of active summaries. Let $\mathbf{m}_n\in\{0,1\}^D$ denote the node-local hard mask and let $\|\mathbf{m}_n\|_0\le K$. The branch logits are
\begin{equation}
\mathbf{g}_n(\mathbf{x}) = r_n(\mathbf{x}\odot \mathbf{m}_n),
\qquad
\mathbf{p}_n(\mathbf{x})=\mathrm{softmax}(\mathbf{g}_n(\mathbf{x})),
\end{equation}
where $r_n$ is a small local router. In the ablations in Table~\ref{tab:training_loss}, $r_n$ can be either a sparse linear map,
\begin{equation}
\mathbf{g}_n(\mathbf{x}) =
W_n(\mathbf{x}\odot \mathbf{m}_n)+\mathbf{b}_n,
\end{equation}
or an additive one-dimensional router,
\begin{equation}
\mathbf{g}_n(\mathbf{x}) =
\sum_{j:m_{n,j}=1} \mathbf{h}_{n,j}(x_j),
\end{equation}
with each $\mathbf{h}_{n,j}$ implemented as a small scalar-to-logit network. These choices are implementation variants for fitting the tree; the final hard MDT exposes the same object in either case: a node-local set of at most $K$ summaries and the resulting branch probabilities.

\subsection{SCCS Sampling and Verification}
\label{app:sccs_details}

SCCS is designed to extract a rule for a held-out target without showing the target policy to the downstream LLM. The sampling procedure first fixes the public context, then looks for private-hand changes that both cross an MDT routing boundary and induce a clear solver-policy change. The resulting prompt displays only the contrastive hands used to form the rule; the evaluation target is required to be policy-separated from every displayed hand.

\begin{algorithm}[ht]
   \caption{SCCS: Scenario-Constrained Contrastive Rule Extraction}
   \label{alg:sccs}
\begin{algorithmic}[1]
   \REQUIRE Target hand $h_{\mathrm{test}}$, trained MDT $\mathcal{M}$, reference solver set $\mathcal{D}$, policy-divergence threshold $\tau_{\pi}$, prompt-separation threshold $\tau_{\mathrm{sep}}$
   \ENSURE SCCS prompt containing contrastive rule evidence, with the target policy masked

   \STATE Fix the public scenario $c=\mathrm{Context}(h_{\mathrm{test}})$.
   \STATE Compute target summaries $\mathbf{x}_{\mathrm{test}}$, target route $P_{\mathrm{test}}=\mathrm{Route}_{\mathcal{M}}(\mathbf{x}_{\mathrm{test}})$, and target policy $\pi^*(\cdot\mid h_{\mathrm{test}})$.
   \STATE Initialize candidate set $\mathcal{C}\leftarrow\emptyset$.

   \STATE \textbf{Scenario-constrained candidate sampling}
   \STATE For each solver-labeled hand $h'\in\mathcal{D}$ with $\mathrm{Context}(h')=c$:
   \STATE \hspace{1em} compute $\mathbf{x}'$, $P'=\mathrm{Route}_{\mathcal{M}}(\mathbf{x}')$, and $\pi^*(\cdot\mid h')$.
   \STATE \hspace{1em} discard $h'$ if $P'=P_{\mathrm{test}}$.
   \STATE \hspace{1em} discard $h'$ if $\bar{L}_1(\pi^*(\cdot\mid h'),\pi^*(\cdot\mid h_{\mathrm{test}}))<\tau_{\pi}$.
   \STATE \hspace{1em} otherwise add $h'$ to $\mathcal{C}$.

   \STATE \textbf{Boundary localization}
   \STATE For each candidate $h'\in\mathcal{C}$, identify the earliest node $n(h')$ where $P'$ and $P_{\mathrm{test}}$ diverge.
   \STATE At $n(h')$, collect the active summaries selected by the hard MDT router.
   \STATE Rank candidates by policy divergence, routing-boundary clarity, and sparsity of the active-summary contrast.
   \STATE Select one or more shadow hands $\mathcal{S}$ from the top-ranked candidates.

   \STATE \textbf{Rule construction}
   \STATE For each shadow hand $h_s\in\mathcal{S}$, extract the diverging node, route branch, active summaries, summary values, MDT policy, and solver policy.
   \STATE Convert these contrasts into a short natural-language rule that names the public scenario, the shadow-hand contrast, and the active summary boundary.

   \STATE \textbf{Held-out verification prompt}
   \STATE Hide $\pi^*(\cdot\mid h_{\mathrm{test}})$ from the downstream LLM.
   \STATE Require $\bar{L}_1(\pi^*(\cdot\mid h_{\mathrm{test}}),\pi^*(\cdot\mid h_s))\ge\tau_{\mathrm{sep}}$ for every displayed hand $h_s\in\mathcal{S}$.
   \STATE Query the LLM with the public scenario, target hand, target summaries, and SCCS rule evidence; evaluate its predicted distribution against the masked target policy.
\end{algorithmic}
\end{algorithm}

Here $\bar{L}_1$ denotes the same action-averaged $L_1$ distance used in the main evaluation. In the communicability experiments, we use $\tau_{\mathrm{sep}}=0.20$ to prevent the displayed SCCS examples from serving as near-label lookups for the held-out target.

\begin{figure*}[ht]
\centering
\begin{small}
\begin{tabular}{p{0.95\linewidth}}
\toprule
\textbf{Input (SCCS Prompt - Abbreviated)} \\
\textbf{Scenario:} BTN vs BB 3-Bet. \textbf{Board:} $4\spadesuit 3\heartsuit 3\diamondsuit$. \textbf{Hand:} $T\heartsuit 8\heartsuit$ (Hero). \\
\textbf{Decision Trace:} Node \#0 (Filter) $\to$ Node \#21 (Split) $\to$ Leaf \#X (Strategy: Bet 100\%). \\
\textbf{SCCS Comparison:} Shadow Hand $K\heartsuit 9\heartsuit$ (Shadow1) diverges at Node \#21 to \textbf{Check}. \\
\textbf{Critical Summary Contrast:}
Hero \texttt{Phys:Kicker} = 0.57 ($T$-High). Shadow1 \texttt{Phys:Kicker} = 0.64 ($K$-High). \\
Hero \texttt{EQ - Nut MDF} = 0.03. Shadow1 \texttt{EQ - Nut MDF} = 0.20. \\
\midrule
\textbf{Model Output (Hard MDT Logic - Generated by Gemini-3.1-Pro)} \\
\textbf{1. Strategy Summary:} The model executes a Polarized Continuation Bet strategy. It aggressively semi-bluffs hands with low Showdown Value but high potential ($T$-high flush draws) to deny equity, while checking back stronger $K$-high hands to realize equity as (S.D. Value).

\textbf{2. Path Logic \& Divergence:}
\begin{itemize}
    \item \textbf{Node \#21 (The Polarization Split):} This is the pivot point separating Semi-Bluffs from S.D. Value.
    \item \textbf{Hero ($T\heartsuit 8\heartsuit$) $\to$ Bet:} Hero has a weaker \texttt{Phys:Kicker} (0.57). The model identifies this hand as having insufficient Showdown Value to check-call. It bets to fold out better hands (like $Q$-high) and deny their equity realization.
    \item \textbf{Shadow1 ($K\heartsuit 9\heartsuit$) $\to$ Check:} Shadow1 possesses a stronger \texttt{Phys:Kicker} (0.64). Despite similar raw equity, the $K$-high acts as a blocker and has enough Showdown Value to beat missed draws. Checking maximizes EV by avoiding isolation against the villain's calling range.
\end{itemize}

\textbf{3. Test Case Solution (Unseen Hand: $K\diamondsuit 9\diamondsuit$)}
\begin{itemize}
    \item \textbf{Predicted Action:} \textbf{Check} (Frequency $\sim$60\%).
    \item \textbf{Reasoning:} The target hand follows the same routed strategic pattern as Shadow1 ($K\heartsuit 9\heartsuit$). With a Kicker of 0.64 ($K$-High), it falls into the S.D. Value bucket. It is too strong to turn into a pure bluff (like the $T\heartsuit 8\heartsuit$) but too weak to value bet, confirming the model's preference for Range Protection over Polarization for this specific kicker class.
\end{itemize} \\
\bottomrule
\end{tabular}
\end{small}
\caption{\textbf{Example of SCCS Rule Extraction.} The SCCS pipeline successfully distills the subtle strategic boundary between a \textit{Semi-Bluff} and \textit{Showdown Value}. The Hard MDT uses the summary \texttt{Phys: Kicker} as a precise cut-off: on a $4\spadesuit 3\heartsuit 3\diamondsuit$ board, $K$-High (Kicker 0.64) is strong enough to Check, whereas $T$-High (Kicker 0.57) must Bet to deny equity. The LLM correctly generalizes this rule to the unseen target hand $K\diamondsuit 9\diamondsuit$.}
\label{fig:llm_example}
\end{figure*}

\section{Detailed Optimization Curriculum}
\label{app:optimization}

Here we detail the three-phase curriculum for the \textbf{Hierarchical Hard Distillation}. This curriculum is designed to distill knowledge from a dense Oracle into a strict symbolic structure without performance collapse.

\paragraph{Phase 1: The Global Teacher (Denoising).}
We first train a ``Global-Gated ResNet'' (Teacher 1). It is restricted to use only $K_g=50$ summaries globally but allows unrestricted non-linear interactions. This phase acts as a filter, removing strictly irrelevant summaries while preserving high-dimensional correlations inherent in the solver data.

\paragraph{Phase 2: The Structural Teacher (Topology).}
We then distill Teacher 1 into a ``Deep Tree'' (Teacher 2). This model adopts the target tree topology but uses higher-capacity local routers. This step establishes the correct decision hierarchy (e.g., branching on Board Texture before Kickers) without being constrained by limited routing capacity.

\paragraph{Phase 3: Hard Student Locking (Logic Extraction).}
Finally, we distill Teacher 2 into the target ``Hard MDT''. Crucially, we switch from soft gating to ``Hard Gating'' using the Straight-Through Estimator (STE). We physically lock the summary set to the Top-$K$ (e.g., 5) at each node:
$$ \mathbf{m}_{hard} = \text{TopK}(\mathbf{m}_{logits}, k=5) $$
This forces the student to find the optimal strategy that exists \textit{strictly within} the 5-summary subspace, mathematically eliminating the residue problem found in soft optimization.

\paragraph{Objective scaling and checkpoint selection.}
In the reported configuration, the distribution and oracle-conditioned EV terms use a $5{:}1$ coefficient ratio. The distribution term is primary: every training phase selects its retained checkpoint by the true validation action-averaged $L_1$. The EV term is an auxiliary safeguard against moving probability toward an action with a large local value cost; it is not used as the primary model-selection metric.

\section{Route-only Component and Mixed-Frequency Diagnostics}
\label{app:route_only}

\paragraph{Component definition.}
Route-only uses the same held-out target, public context, legal actions, and displayed MDT route trace as SCCS. It includes the visited nodes, branches, routing/reach probabilities, node-local summary impacts, and reached pure-action prototype. It removes every reference or shadow hand, every displayed reference policy, and every directional cross-hand rule. The target solver policy and the aggregate MDT mixed policy remain hidden in all prompting conditions. Route-only therefore asks whether the tree trace alone is sufficient; SCCS adds the feasible same-context comparison and the policy change across the learned boundary.

This control is also the closest measured comparison to several generic explanation alternatives. Feature attribution can identify influential quantities for one prediction, but does not by itself construct a feasible private state under the same public history, verify that its solver mixture differs, or express the direction in which probability moves across the learned boundary. Nearest-neighbor retrieval need not preserve the public scenario or cross a policy-relevant route boundary, while a human heuristic or compressed strategy table need not be tied to the summaries that actually determine MDT's computation. Direct+Summaries tests whether the unstructured numerical inputs alone are sufficient, and Route-only tests whether the displayed hierarchy alone is sufficient. Our experiments therefore isolate the incremental value of the matched shadow-based contrast over these two implemented controls; they do not claim a comparison with unimplemented attribution, retrieval, or hand-authored systems.

Table~\ref{tab:route_aggregation} distinguishes the two aggregation conventions used in our analyses. The main Table~\ref{tab:comm} follows the submitted convention, in which each prompt condition is averaged over its own parse-success set and the eight configuration means are then averaged without weighting. The paired comparison instead restricts Route-only and SCCS to the same parsed target IDs within each model configuration.

\begin{table}[ht]
\caption{\textbf{Route-only versus SCCS under two aggregation conventions.} Values are action-averaged $L_1$; parenthesized values are relative reductions from Route-only to SCCS.}
\label{tab:route_aggregation}
\centering
\begin{small}
\begin{tabular}{lcc}
\toprule
\textbf{Aggregation} & \textbf{$L_1$ to solver} & \textbf{$L_1$ to MDT} \\
\midrule
Condition-wise Table~\ref{tab:comm} means & $0.173\rightarrow0.100$ ($42.2\%$) & $0.172\rightarrow0.114$ ($33.7\%$) \\
Shared-parse paired means & $0.170\rightarrow0.100$ ($41.2\%$) & $0.172\rightarrow0.114$ ($33.8\%$) \\
\bottomrule
\end{tabular}
\end{small}
\end{table}

Route-only has 100 parsed targets in seven configurations and 99 in Gemini-3.1 Flash. The retained SCCS outputs parse on 100, 100, 15, 100, 99, 68, 100, and 29 targets for Gemini-3.1 Flash, Gemini-3.1 Pro low/high, DeepSeek-V4 Flash/Pro, and GPT-5.4/5.5 low/high, respectively. The main table reports these parse-conditional condition means; the paired diagnostics never compare different target IDs. For the three GPT configurations, the Route-only calls used a separate API wrapper from the saved SCCS calls while retaining the same named model and reasoning setting.

\paragraph{Holding the dominant action fixed.}
We perform two post-hoc, parse-conditional diagnostics. First, among 342 matched model--target cases where Direct already selects the solver argmax, Route-only and SCCS have nearly identical argmax agreement ($81.3\%$ and $80.1\%$), but SCCS lowers Route-only solver $L_1$ from $0.15633$ to $0.10082$, a $35.5\%$ reduction. SCCS also lowers Direct's remaining solver $L_1$ from $0.11948$ to $0.10082$, a $15.6\%$ reduction.

Second, we restrict further to the 230 model--target cases where Direct, Route-only, and SCCS all select the correct solver argmax. SCCS lowers Route-only solver $L_1$ from $0.12506$ to $0.07648$ ($38.8\%$) and MDT $L_1$ from $0.14908$ to $0.09984$ ($33.0\%$). The configuration-level point estimate favors SCCS in all eight configurations under both targets. Because dominant-action correctness is fixed, these differences measure the remaining probability allocation rather than an argmax-action correction. They are diagnostic subset analyses, not a formal probability-calibration study, and repeated target states across model configurations are not treated as independent game samples.

\section{Liar's Dice Experimental Details}
\label{app:liars_dice}

\paragraph{Game and protocol.}
The game has two players, two three-faced dice per player, and bids up to quantity three. Because a player's two dice are unordered and repetitions are allowed, there are six private types: $11,12,13,22,23,33$. A full-tree CFR+ solver produces the approximate-Nash policy. The MDT interface contains 17 public-history, posterior-belief, private-state, and equity summaries; per-action continuation Q-values, Q-gaps, and the target policy are excluded from both MDT inputs and model-facing prompts. Six cross-fitting folds each hold out one private type at every public context. The communication surface contains 31 held-out target information states and four matched prompts per target.

This 31-target surface is a post-result exploratory expansion of an earlier pilot. Its eligibility and anti-copy filters, prompts, and admissible reference/shadow ranking were frozen before collecting outputs on the expanded surface. Metrics are computed over successfully parsed responses, and unavailable responses are not imputed. Table~\ref{tab:liars_dice_full} reports the complete released configuration-level results.

\begin{table}[ht]
\caption{\textbf{Per-configuration Liar's Dice communicability.} Model entries are mean $\pm$ standard error over parse-success cases. Asterisks mark incomplete provider runs; missing responses are not imputed.}
\label{tab:liars_dice_full}
\centering
\begin{scriptsize}
\resizebox{\textwidth}{!}{\begin{tabular}{lcccc|cccc}
\toprule
\multirow{2}{*}{\textbf{LLM run}} & \multicolumn{4}{c|}{\textbf{$L_1$ to solver}} & \multicolumn{4}{c}{\textbf{$L_1$ to MDT}}\\
& Direct & +Summaries & +Route-only & +SCCS & Direct & +Summaries & +Route-only & +SCCS\\
\midrule
Gemini-3.1 Flash & $.141\pm.019$ & $.165\pm.022$ & $.107\pm.012$ & $.096\pm.015$ & $.154\pm.023$ & $.167\pm.026$ & $.087\pm.010$ & $.068\pm.009$ \\
Gemini-3.1 Pro low & $.171\pm.020$ & $.149\pm.017$ & $.132\pm.014$ & $.123\pm.011$ & $.192\pm.026$ & $.167\pm.023$ & $.095\pm.013$ & $.078\pm.009$ \\
Gemini-3.1 Pro high$^*$ & $.168\pm.019$ & $.148\pm.014$ & $.125\pm.011$ & $.100\pm.011$ & $.184\pm.025$ & $.148\pm.020$ & $.090\pm.009$ & $.080\pm.010$ \\
DeepSeek-V4 Flash & $.266\pm.015$ & $.270\pm.014$ & $.134\pm.012$ & $.120\pm.012$ & $.271\pm.019$ & $.273\pm.018$ & $.089\pm.009$ & $.072\pm.009$ \\
DeepSeek-V4 Pro & $.280\pm.012$ & $.280\pm.012$ & $.128\pm.014$ & $.111\pm.013$ & $.285\pm.016$ & $.285\pm.016$ & $.089\pm.009$ & $.088\pm.011$ \\
GPT-5.4 & $.113\pm.013$ & $.120\pm.012$ & $.094\pm.012$ & $.088\pm.012$ & $.124\pm.017$ & $.126\pm.017$ & $.072\pm.006$ & $.077\pm.008$ \\
GPT-5.5 low & $.112\pm.014$ & $.113\pm.017$ & $.101\pm.013$ & $.094\pm.011$ & $.131\pm.018$ & $.104\pm.015$ & $.068\pm.006$ & $.083\pm.008$ \\
GPT-5.5 high$^*$ & $.137\pm.018$ & $.124\pm.015$ & $.105\pm.013$ & $.104\pm.012$ & $.153\pm.024$ & $.120\pm.016$ & $.068\pm.007$ & $.086\pm.008$ \\
\midrule
Mean & $.174\pm.065$ & $.171\pm.067$ & $.116\pm.016$ & $\mathbf{.105\pm.012}$ & $.187\pm.061$ & $.174\pm.069$ & $.082\pm.011$ & $\mathbf{.079\pm.007}$ \\
\bottomrule
\end{tabular}}
\end{scriptsize}
\end{table}

Six configurations contain all 124 expected responses. Gemini-3.1 Pro high contains 117 of 124, and GPT-5.5 high contains 123 of 124. In the order Direct, Summaries, Route-only, and SCCS, the parse counts out of 31 are: Gemini-3.1 Flash $(24,27,31,15)$; Gemini-3.1 Pro low $(31,31,31,31)$; Gemini-3.1 Pro high $(29,28,31,29)$; DeepSeek-V4 Flash $(30,31,31,30)$; DeepSeek-V4 Pro $(31,31,30,29)$; GPT-5.4 and GPT-5.5 low $(31,31,31,31)$; and GPT-5.5 high $(30,31,31,31)$. The unweighted Mean row therefore summarizes configuration-level, parse-conditional point estimates rather than eight complete replications. On paired parse-success targets, the expanded-surface point estimate favors SCCS over Direct and Direct+Summaries in every listed configuration, whereas the smaller SCCS--Route-only margins are heterogeneous and should be interpreted as a component diagnostic.

\paragraph{Complete fixed-opponent value evaluation.}
The six held-out folds provide exactly one MDT prediction for each of the game's 3,072 information sets. We assemble these predictions into a fixed behavioral policy and traverse the complete tree against the approximate-Nash solver, once in each player position. Relative to solver self-play, the MDT value losses are $0.080723$ as player 0 and $0.035129$ as player 1, for a seat average of $0.057926$ (reported as $0.058$) per game on the terminal payoff scale $[-1,+1]$. This measures the complete policy's value against a fixed reference opponent; it does not recompute an adaptive best response.

\section{Complete River-Endgame Evaluation Details}
\label{app:river_eval}

\paragraph{Public states and action trees.}
The experiment starts from the released HUNL River states used for Subgames 3 and 4 by \citet{brown2019solving}. Subgame 3 uses board \texttt{4s 8h Tc 9h 2s}, a $5$ bb root pot, and $197.5$ bb remaining per player. Its two root bet sizes are $3.4/7.5$ bb and its two bet sizes after a check are $2/3.4$ bb. Subgame 4 uses board \texttt{Js Ks 5c Qs 7d}, a $37.5$ bb root pot, and $181.25$ bb remaining per player. Its corresponding sizes are $9.4/37.5$ bb at the root and $56.3/112.5$ bb after a check. We retain the released root ranges and build complete local trees through fold or showdown. These are our own two-size trees initialized from the published states, not the betting abstraction used in the original benchmark curves.

Local DCFR solves provide high-precision supervised targets. Their residual exact exploitabilities are $0.009034$ mbb/g on Subgame 3 and $0.009907$ mbb/g on Subgame 4. Both MDT variants use depth-4 ternary probabilistic routing and pure-action leaves, with one model per acting position shared across that position's decision states. The inputs exclude action Q/value features, target-policy probabilities, and exact private-card indicators. Soft can use all eligible summaries; Hard uses exactly five summaries at each router. After supervised fitting, the deployments are fine-tuned and selected using the exact same-tree strategic metric. The result is therefore an in-domain strategic-fit evaluation of the compressed policies.

\begin{table}[ht]
\caption{\textbf{Additional River-endgame diagnostics.} $L_1$ is equally averaged over decision states. Fixed-target value loss keeps the local target opponent fixed; exact exploitability recomputes each player's best response. All value entries are mbb/g.}
\label{tab:river_details}
\centering
\begin{small}
\begin{tabular}{llrrr}
\toprule
\textbf{State} & \textbf{Policy} & \textbf{$L_1$} & \textbf{Fixed-target loss} & \textbf{Exact exploitability} \\
\midrule
Subgame 3 & Soft MDT & $0.01214$ & $1.909$ & $6.691$ \\
& Hard Top-5 MDT & $0.03751$ & $7.704$ & $17.173$ \\
\midrule
Subgame 4 & Soft MDT & $0.02746$ & $25.917$ & $55.013$ \\
& Hard Top-5 MDT & $0.05964$ & $54.951$ & $133.046$ \\
\bottomrule
\end{tabular}
\end{small}
\end{table}

For the matched learning-based reference, Deep CFR uses $10{,}000$ sampled game-tree traversals per player per iteration. We evaluate its linearly weighted empirical-average strategy, rather than fitting a separate final actor, with the same exact best-response computation. At iteration 800 it reaches $43.075$ mbb/g on Subgame 3 and $133.495$ mbb/g on Subgame 4. These checkpoints give scale on the same trees and metric; they are neither convergence limits nor a general comparison between MDT and Deep CFR outside these two in-domain endgames.

\section{Metric Interpretation and Fixed-Opponent NLH Evaluation}
\label{app:metric_interpretation}

\paragraph{$L_1$ is primary and local EV is auxiliary.}
The action-averaged $L_1$ directly measures fidelity to the solver's complete local mixture. Its unaveraged form is twice total variation, so probability-mass displacement changes the metric linearly. By contrast, the oracle-conditioned EV gap evaluates the predicted mixture using fixed solver continuation values. At an exact equilibrium information set, every supported solver action is value-maximizing; under that condition, the one-decision EV gap equals the predicted mixture's one-step regret relative to the best action. Our targets are approximate solver solutions, so we use the more limited term \emph{local value gap}. It is not CFR counterfactual regret or a complete-policy best-response metric.

EV alone is insufficient for the articulation objective. Multiple supported equilibrium actions can be nearly indifferent against the fixed equilibrium opponent, so an almost-pure prediction can have low local EV loss while remaining far from the intended mixture. $L_1$ therefore determines checkpoint selection, while the EV term discourages locally costly probability shifts among predictions with similar distributional fidelity.

\paragraph{Corpus-scale fixed-opponent comparison.}
As a complementary NLH diagnostic, we evaluate Soft-sparse Tree and final Hard MDT against the fixed approximate-Nash opponent on the solver-labeled postflop distribution. Their value losses are $7.836$ and $20.688$ mbb per evaluated postflop hand, respectively, a descriptive difference of $12.852$ mbb per evaluated postflop hand. This evaluation keeps the opponent and solver-derived summaries fixed. The checkpoint families also differ in more than the Top-5 mask, so the controlled same-architecture River results in Section~\ref{sec:river_eval} provide the cleaner measurement of strict-sparsity cost.

\section{Limitations and Future Work}

While our framework successfully distills articulate reasoning from solver data, we identify several current limitations regarding its scope and deployment.

\paragraph{Dependence on Mixed-Strategy Equilibria}
The Mixed-Strategy Decision Tree (MDT) is architecturally specialized for imperfect-information games characterized by mixed Nash equilibria. The model's inductive bias, specifically its decomposition of strategy into probabilistic routing over pure-action leaf prototypes, is designed to capture the delicate frequency balancing required in games like No-Limit Texas Hold'em. Consequently, this approach may yield diminishing returns in perfect-information domains (e.g., chess) or games dominated by pure strategies, where such complex mixing specifications are unnecessary.

\paragraph{Dependence on Solver-Derived Summaries}
Currently, our system operates as an offline analytical agent rather than a standalone poker agent. Because the MDT relies on solver-derived strategic summaries, it cannot yet function in a live setting where such ground-truth solver information is unavailable. Bridging this gap, potentially by training a separate state-estimation module to approximate these summaries from raw history, remains a critical direction for future work to enable live-agent deployment.

\paragraph{Licensed Data and Reproducibility}
The commercial NLH solver corpus cannot be redistributed in bulk. We can release the solver-interface schema, MDT/SCCS implementation, configuration, prompt and evaluation code, aggregate results, and legally distributable examples, but exact regeneration of the 250-million-decision corpus requires access to a compatible licensed solver. The Liar's Dice experiment addresses this reproducibility boundary with a releasable game solver, data generator, solver targets, trained cross-fitted MDTs, prompts, outputs, and scoring code.

\paragraph{Action Abstraction and Coverage}
The NLH targets use response sizes selected by the solver abstraction. A non-standard earlier bet size can be represented as part of the observed history and changes the state summarized for MDT, but the current prediction remains a mixture over the solver-supported response actions. We do not claim robustness to arbitrary histories outside this interface. SCCS also requires a feasible same-context shadow satisfying both route- and policy-divergence constraints. If no such shadow exists, the system can expose the Route-only trace but cannot construct a supported contrastive rule.

\paragraph{Inference-Time Evaluation}
The downstream LLM parameters are fixed in all reported communicability experiments. The demonstrated result is that an independent LLM can use solver-derived rules at inference time to improve prediction of a hidden mixed policy. Training language models on a large corpus of such rules is a natural extension, but is not evaluated here.

\section{Prompt Templates}
\label{app:prompt}

We use a structured prompt to ground the LLM's generation in the routed logic exposed by MDT. The template below is populated dynamically by the SCCS engine. We also provide example output of Gemini-3.1-Pro.

Template 1:
\begin{promptlisting}
# Role
You are an elite Poker AI Strategist and GTO Solver Analyst. Your task is to reverse-engineer the decision-making process exposed by a Mixed-Strategy Decision Tree (MDT) model.

# Task
Analyze the provided trace logic for the specific hand provided below. You must explain *why* the AI chose this specific path over others, using Poker Theory concepts (Equity Realization, Blockers, Range Morphology).

# Semantic Definitions (Crucial)
The trace uses specific tags based on Equity (EQ) vs. Opponent Range:
- **(Value)**: Aggressive action with High EQ (>65%).
- **(Bluff)**: Aggressive action with Low EQ (<35%).
- **(Thin Value/Protect)**: Aggressive action with Moderate EQ.
- **(S.D. Value)**: Passive action (Check/Call) with enough EQ to win at showdown but not enough to bet for value.
- **(Give Up)**: Passive action with near-zero EQ.
- **(Trap)**: Passive action with Nut-class EQ.

# Input Data
=== SCENARIO CONTEXT ===
Scenario : BTN_vs_BB_3B
Board    : 4s3h3d
History  : k
Player   : IP
Hand     : Th8h
Hand Wgt : 0.0039 (Prob in Range)
Pot      : 26.5
Actions  : [Check, Fold, Bet(0.48x)]

=== SUMMARY GLOSSARY (Definitions for current path) ===

--- Strategic Adv ---
- **Hero EQ Adv** (Avg: -0.03): **RELATIVE SUMMARY (Hero - Villain)**. Difference in Range Equity. Positive (>0) = Hero is Ahead. Negative (<0) = Villain is Ahead. (Automatically adjusted from Global OOP-IP value based on Hero's position).
- **Hero EV Adv** (Avg: -0.19): **RELATIVE SUMMARY (Hero - Villain)**. Difference in EV Pot Share. Positive = Favors Hero.
- **Hand EQ** (Avg: 0.49): Raw Equity (0.0-1.0) against opponent's current range.
- **EQ - Nut MDF** (Avg: 0.25): Hand Equity minus Nut MDF. Positive = Hand is strong enough to play for stacks.
- **EQ - Nut Range Avg** (Avg: 0.27): Hand EQ minus the Average EQ of the Hero's Nut Range. Are you Top Nut or Bottom Nut?
- **Rank in Range (EQ)** (Avg: 0.50): Percentile of Hand Equity (0-1).

--- Hand Physics ---
- **Phys: Kicker** (Avg: 0.65): Normalized Kicker Strength (Rank / 14.0). Ace=1.0, King=0.92, ..., 2=0.14. Crucial for domination issues (e.g., distinguishing Top Pair Top Kicker from Top Pair Weak Kicker).

--- Range Buckets ---
- **OOP Rng EV** (Avg: 0.59): OOP's EV expressed as a percentage of the total pot.

--- Other ---
- **Bluff Efficiency** (Avg: 0.61): Measure of how well our 'Air' range blocks opponent's calling range. High Efficiency = We have 'Natural Bluffs' (Low Equity + High Blocker Score). Low Efficiency = Our bluffs have poor removal (random trash).
- **Hand Blocker** (Avg: 3.62): Score representing how much this hand blocks opponent's continuing range.
- **Hand EV** (Avg: 0.41): Expected Value of the hand normalized by Pot.
- **MDF (Decision)** (Avg: 0.65): Decision-based MDF = Pot / (Pot + Bet). The break-even equity required to call.
- **Vuln - Range Avg** (Avg: 0.00): Hand Vulnerability minus Range Average. Positive = More vulnerable than average (needs protection).
- **Regret: Fold-Call** (Avg: 0.89): EV(Fold) - EV(Call). Positive = Fold is better. Diff in EV (Action A - Action B) normalized by Pot.

=== DECISION LOGIC TRACE ===
Legend:
 - [Path]: format is 'Role [Hand] -> PathID -> IntendedAction'. Indicates which internal branch was taken and the final action intent.
 - (Reach): The probability of the hand actually reaching this node vs. surviving to the next node.
 - R: [0:xx 1:xx ...]: Router Probabilities. The internal neural network's confidence distribution across Branch 0, 1, and 2. Shows how 'split' or 'certain' the decision was.
 - [Summary Impact]: Shows which summaries pushed the router towards specific branches.
+-- NODE #0
      [Path] Hero    [Th8h] -> P0->Bet    (Reach: 100%->96%) | R: [0:95 1:00 2:04]
      [Path]  Shadow1 [Kh9h] -> P0->Check  (Reach: 100%->74%) | R: [0:73 1:00 2:26]
      [Path]  Shadow2 [Tc9c] -> P0->Check  (Reach: 100%->100%) | R: [0:99 1:00 2:00]
      [Path]  Shadow3 [Ks6s] -> P0->Check  (Reach: 100%->63%) | R: [0:62 1:00 2:37]
      [Summary Impact Analysis - Node #0]
      | Summary                   | Avg      | Hero Impact (Chk-Bet)               | Shadow1      | Shadow2      | Shadow3      |
      | :------------------------ | :------: | :---------------------------------: | :----------: | :----------: | :----------: |
      | Rank in Range (EQ)        | 0.50     | **0.06** (+29.1)                    | 0.24         | 0.02         | 0.25         |
      | Hand EQ                   | 0.49     | **0.26** (+15.1)                    | 0.43         | 0.22         | 0.44         |
      | MDF (Decision)            | 0.65     | **-1.00** (-6.3)                    | -1.00        | -1.00        | -1.00        |
      | Phys: Kicker              | 0.65     | **0.57** (-2.1)                     | 0.64         | 0.64         | 0.43         |
      | Hand EV                   | 0.41     | **0.32** (-0.1)                     | 0.34         | 0.24         | 0.38         |
    +-- NODE #1
          [Path] Hero    [Th8h] -> P2->Bet    (Reach: 96%->75%) | R: [0:24 1:00 2:75]
          [Path]  Shadow1 [Kh9h] -> P2->Check  (Reach: 74%->100%) | R: [0:00 1:00 2:99]
          [Path]  Shadow2 [Tc9c] -> P2->Check  (Reach: 100%->62%) | R: [0:37 1:00 2:62]
          [Path]  Shadow3 [Ks6s] -> P2->Check  (Reach: 63%->100%) | R: [0:00 1:00 2:99]
          [Summary Impact Analysis - Node #1]
          | Summary                   | Avg      | Hero Impact (Bet-Chk)               | Shadow1      | Shadow2      | Shadow3      |
          | :------------------------ | :------: | :---------------------------------: | :----------: | :----------: | :----------: |
          | Hand EV                   | 0.41     | **0.32** (+29.1)                    | 0.34         | 0.24         | 0.38         |
          | Hero EV Adv               | -0.19    | **0.12** (-27.5)                    | 0.12         | 0.12         | 0.12         |
          | Vuln - Range Avg          | 0.00     | **-0.16** (+15.8)                   | -0.04        | -0.13        | -0.05        |
          | Hand Blocker              | 3.62     | **1.00** (+1.7)                     | 3.00         | 2.00         | 3.00         |
          | Hand EQ                   | 0.49     | **0.26** (-0.0)                     | 0.43         | 0.22         | 0.44         |
        +-- NODE #6
              [Path] Hero    [Th8h] -> P2->Bet    (Reach: 72%->100%) | R: [0:00 1:00 2:99]
              [Path]  Shadow1 [Kh9h] -> P2->Check  (Reach: 74%->100%) | R: [0:00 1:00 2:99]
              [Path]  Shadow2 [Tc9c] -> P2->Check  (Reach: 62%->100%) | R: [0:00 1:00 2:99]
              [Path]  Shadow3 [Ks6s] -> P2->Check  (Reach: 63%->100%) | R: [0:00 1:00 2:99]
              [Summary Impact Analysis - Node #6]
              | Summary                   | Avg      | Hero Impact (Bet-(Chk))             | Shadow1      | Shadow2      | Shadow3      |
              | :------------------------ | :------: | :---------------------------------: | :----------: | :----------: | :----------: |
              | Hand EV                   | 0.41     | **0.32** (+51.2)                    | 0.34         | 0.24         | 0.38         |
              | Hero EQ Adv               | -0.03    | **0.09** (-20.7)                    | 0.09         | 0.09         | 0.09         |
              | Phys: Kicker              | 0.65     | **0.57** (+7.4)                     | 0.64         | 0.64         | 0.43         |
              | Bluff Efficiency          | 0.61     | **0.42** (+7.2)                     | 0.42         | 0.42         | 0.42         |
              | EQ - Nut Range Avg        | 0.27     | **0.03** (+4.7)                     | 0.20         | -0.02        | 0.20         |
            +-- NODE #21
                  [Path] Hero    [Th8h] -> P2->Bet    (Reach: 72%->100%) | R: [0:00 1:00 2:99]
                  [Path]  Shadow1 [Kh9h] -> P0->Check  (Reach: 74%->81%) | R: [0:81 1:00 2:18]
                  [Path]  Shadow2 [Tc9c] -> P2->Check  (Reach: 62%->95%) | R: [0:02 1:02 2:95]
                  [Path]  Shadow3 [Ks6s] -> P2->Check  (Reach: 63%->97%) | R: [0:02 1:00 2:97]
                  [Summary Impact Analysis - Node #21]
                  | Summary                   | Avg      | Hero Impact (Bet-(Chk))             | Shadow1      | Shadow2      | Shadow3      |
                  | :------------------------ | :------: | :---------------------------------: | :----------: | :----------: | :----------: |
                  | EQ - Nut MDF              | 0.25     | **0.03** (+5.6)                     | 0.20         | -0.01        | 0.20         |
                  | Vuln - Range Avg          | 0.00     | **-0.16** (+4.0)                    | -0.04        | -0.13        | -0.05        |
                  | Phys: Kicker              | 0.65     | **0.57** (+1.6)                     | 0.64         | 0.64         | 0.43         |
                  | Regret: Fold-Call         | 0.89     | **1.02** (+1.0)                     | 0.99         | 1.12         | 0.90         |
                  | OOP Rng EV                | 0.59     | **0.44** (+0.3)                     | 0.44         | 0.44         | 0.44         |
                +-- LEAF #64 (Static Prototype)
                |     -> Strat: [Check:99%] | Samples: [Shadow1:Kh9h] (Reach:59.6%)
                +-- LEAF #66 (Static Prototype)
                      -> Strat: [Check:99%] | Samples: [Hero:Th8h] (Reach:71.8%), [Shadow2:Tc9c] (Reach:58.8%), [Shadow3:Ks6s] (Reach:60.8%)

=== FINAL STRATEGY vs GTO COMPARISON (Training Data) ===
| Hand     | Role     | Final Model Strat    | GTO Target Strat     |
| :------: | :------: | :------------------: | :------------------: |
| Th8h     | Hero     | C:0.40 F:0.00 B:0.60 | C:0.40 F:0.00 B:0.60 |
| Kh9h     | Shadow1  | C:0.61 F:0.00 B:0.39 | C:0.59 F:0.00 B:0.41 |
| Tc9c     | Shadow2  | C:0.66 F:0.00 B:0.34 | C:0.35 F:0.00 B:0.65 |
| Ks6s     | Shadow3  | C:0.52 F:0.00 B:0.48 | C:0.41 F:0.00 B:0.59 |

## 3. Test Case: The Unseen Hand
Consider a new hand in the same scenario:
- **Hand**: Kd9d
- **Key Difference**: (Inspect the summaries yourself compared to Hero)
**Question**: Based on the logic learned above, what is the optimal action for this hand? Explain why using the model's decision boundaries.
# Analysis Instructions
1. **Layer-by-Layer Review**: For each NODE, explain how the specific **Impact values** determined the branch choice.
2. **Impact Comparison**: Compare Hero's summary values vs. Global Avg and Shadows.
3. **Consistency Check**: Use Section 3 (Test Case) to verify if the logic you reverse-engineered applies to an unseen hand.
4. **Solve Test Case**: Provide your answer for the Unseen Hand in Section 3.
    - Key Summary Values of Test Case (Hierarchical):
    --- **(Depth 0)** - Top 5 Drivers:
      - **Rank in Range (EQ)**: 0.24 (Avg: 0.50) | Impact: +18.28
      - **MDF (Decision)**: -1.00 (Avg: 0.65) | Impact: -6.30
      - **Phys: Kicker**: 0.64 (Avg: 0.65) | Impact: -0.95
      - **Hand EV**: 0.34 (Avg: 0.41) | Impact: -0.50
      - **Hand EQ**: 0.43 (Avg: 0.49) | Impact: +0.20
    --- **(Depth 1)** - Top 4 Drivers:
      - **Hand EV**: 0.34 (Avg: 0.41) | Impact: +31.56
      - **Hero EV Adv**: 0.12 (Avg: -0.19) | Impact: -27.52
      - **Vuln - Range Avg**: -0.04 (Avg: 0.00) | Impact: +13.43
      - **Hand Blocker**: 3.00 (Avg: 3.62) | Impact: +12.95
    --- **(Depth 2)** - Top 3 Drivers:
      - **Hand EV**: 0.34 (Avg: 0.41) | Impact: +55.78
      - **Hero EQ Adv**: 0.09 (Avg: -0.03) | Impact: -20.67
      - **Bluff Efficiency**: 0.42 (Avg: 0.61) | Impact: +7.21
    --- **(Depth 3)** - Top 2 Drivers:
      - **Phys: Kicker**: 0.64 (Avg: 0.65) | Impact: +3.12
      - **EQ - Nut MDF**: 0.20 (Avg: 0.25) | Impact: +1.43

# Output Format (Markdown)
## 1. Strategy Summary
(One concise sentence)

## 2. Path Logic & Divergence
- **Node #0**: ...

## 3. Test Case Solution
- **Predicted Mixed Strategy**: ...
- **Reasoning**: ...
\end{promptlisting}

Answer 1:
\begin{promptlisting}
# Test Case Solution (Sample 5181)
Hand: Kd9d
Scenario: BTN_vs_BB_3B

## 1. Ground Truth Strategies
- **GTO Strategy**:   C:0.59 F:0.00 B:0.41
- **Model Strategy**: C:0.61 F:0.00 B:0.39
## 2. Actual Decision Trace
+-- NODE #0
      [Path]  TestHand [Kd9d] -> P0->Check  (Reach: 100%->73%) | R: [0:73 1:00 2:26]
      [Summary Impact Analysis - Node #0 (Analysis Focus: TestHand - Hero diverged)]
      | Summary                   | Avg      | TestHand Impact (Chk-Bet)           |
      | :------------------------ | :------: | :---------------------------------: |
      | Rank in Range (EQ)        | 0.50     | **0.24** (+18.3)                    |
      | MDF (Decision)            | 0.65     | **-1.00** (-6.3)                    |
      | Phys: Kicker              | 0.65     | **0.64** (-0.9)                     |
      | Hand EV                   | 0.41     | **0.34** (-0.5)                     |
      | Hand EQ                   | 0.49     | **0.43** (+0.2)                     |
    +-- NODE #1
          [Path]  TestHand [Kd9d] -> P2->Check  (Reach: 73%->100%) | R: [0:00 1:00 2:99]
          [Summary Impact Analysis - Node #1 (Analysis Focus: TestHand - Hero diverged)]
          | Summary                   | Avg      | TestHand Impact (Bet-(Chk))         |
          | :------------------------ | :------: | :---------------------------------: |
          | Hand EV                   | 0.41     | **0.34** (+31.6)                    |
          | Hero EV Adv               | -0.19    | **0.12** (-27.5)                    |
          | Vuln - Range Avg          | 0.00     | **-0.04** (+13.4)                   |
          | Hand Blocker              | 3.62     | **3.00** (+12.9)                    |
          | Hand EQ                   | 0.49     | **0.43** (+4.4)                     |
        +-- NODE #6
              [Path]  TestHand [Kd9d] -> P2->Check  (Reach: 73%->100%) | R: [0:00 1:00 2:99]
              [Summary Impact Analysis - Node #6 (Analysis Focus: TestHand - Hero diverged)]
              | Summary                   | Avg      | TestHand Impact (Bet-(Chk))         |
              | :------------------------ | :------: | :---------------------------------: |
              | Hand EV                   | 0.41     | **0.34** (+55.8)                    |
              | Hero EQ Adv               | -0.03    | **0.09** (-20.7)                    |
              | Bluff Efficiency          | 0.61     | **0.42** (+7.2)                     |
              | Phys: Kicker              | 0.65     | **0.64** (+7.1)                     |
              | EQ - Nut Range Avg        | 0.27     | **0.20** (+6.4)                     |
            +-- NODE #21
                  [Path]  TestHand [Kd9d] -> P0->Check  (Reach: 73%->81%) | R: [0:81 1:00 2:18]
                  [Summary Impact Analysis - Node #21 (Analysis Focus: TestHand - Hero diverged)]
                  | Summary                   | Avg      | TestHand Impact (Chk-Bet)           |
                  | :------------------------ | :------: | :---------------------------------: |
                  | Phys: Kicker              | 0.65     | **0.64** (+3.1)                     |
                  | EQ - Nut MDF              | 0.25     | **0.20** (+1.4)                     |
                  | Regret: Fold-Call         | 0.89     | **0.99** (-1.1)                     |
                  | Vuln - Range Avg          | 0.00     | **-0.04** (-0.9)                    |
                  | OOP Rng EV                | 0.59     | **0.44** (-0.3)                     |
                +-- LEAF #64 (Static Prototype)
                      -> Strat: [Check:99%] | Samples: [TestHand:Kd9d] (Reach:59.1%)

## 3. Key Summary Drivers (Path-Dependent)

### **(Depth 0)** - Top 5 Drivers:
- **Rank in Range (EQ)**: 0.24 (Avg: 0.50) | Impact: +18.28
- **MDF (Decision)**: -1.00 (Avg: 0.65) | Impact: -6.30
- **Phys: Kicker**: 0.64 (Avg: 0.65) | Impact: -0.95
- **Hand EV**: 0.34 (Avg: 0.41) | Impact: -0.50
- **Hand EQ**: 0.43 (Avg: 0.49) | Impact: +0.20

### **(Depth 1)** - Top 4 Drivers:
- **Hand EV**: 0.34 (Avg: 0.41) | Impact: +31.56
- **Hero EV Adv**: 0.12 (Avg: -0.19) | Impact: -27.52
- **Vuln - Range Avg**: -0.04 (Avg: 0.00) | Impact: +13.43
- **Hand Blocker**: 3.00 (Avg: 3.62) | Impact: +12.95

### **(Depth 2)** - Top 3 Drivers:
- **Hand EV**: 0.34 (Avg: 0.41) | Impact: +55.78
- **Hero EQ Adv**: 0.09 (Avg: -0.03) | Impact: -20.67
- **Bluff Efficiency**: 0.42 (Avg: 0.61) | Impact: +7.21

### **(Depth 3)** - Top 2 Drivers:
- **Phys: Kicker**: 0.64 (Avg: 0.65) | Impact: +3.12
- **EQ - Nut MDF**: 0.20 (Avg: 0.25) | Impact: +1.43

\end{promptlisting}

LLM Response 1:
\begin{promptlisting}
## 1. Strategy Summary
The model executes a **Polarized Continuation Bet** strategy where hands with low Showdown Value but high potential (Ten-high flush draws) aggressively semi-bluff to deny equity, while stronger King-high hands (S.D. Value) check back to realize equity and protect the checking range.

## 2. Path Logic & Divergence

### Node #0 (The Equity Filter)
*   **Role**: Preliminary Router.
*   **Logic**: This node acts as a low-frequency filter rather than a strategic decision maker. While `Rank in Range` is the primary summary (Hero Impact +29.1), the router sends the vast majority of all analyzed hands (Hero 96%, Shadows 74-100%) down the same pipe (Path 0 -> Node #1).
*   **Divergence**: There is no meaningful divergence here; the model has determined that all these IP hands (Th8h, Kh9h, etc.) belong in the active decision tree rather than an immediate "Give Up" bucket.

### Node #21 (The Polarization Split)
*   **Role**: The primary strategic pivot point separating **Semi-Bluffs** from **S.D. Value**.
*   **Logic**: This node splits Hero (Th8h) from Shadow1 (Kh9h).
    *   **Hero (Th8h) -> Bet**: Hero has a weaker `Phys: Kicker` (0.57) and very low `EQ - Nut MDF` (0.03). The model identifies this as a hand with insufficient Showdown Value to check-call but excellent "Natural Bluff" properties (Backdoor Flush Draw + Overcards). It bets to fold out better hands (Q-high, J-high) and deny their equity realization.
    *   **Shadow1 (Kh9h) -> Check**: Shadow1 has a stronger `Phys: Kicker` (0.64) and higher `EQ - Nut MDF` (0.20). Despite having similar raw equity, the King-high acts as a blocker to the opponent's continuing range and has enough Showdown Value to beat missed draws or lower high-cards. Checking maximizes EV by avoiding isolation against the villain's calling range (which is pair-heavy).

## 3. Test Case Solution

*   **Hand**: **Kd9d** (King-Nine of Diamonds)
*   **Predicted Mixed Strategy**: **Check** (Frequency ~60% Check / 40% Bet)
*   **Reasoning**:
    1.  **Morphological Twin**: Kd9d is effectively identical to Shadow1 (Kh9h). Both are King-high hands with a backdoor flush draw on a paired low board (4s3h3d). In MDT models, identical summary vectors result in identical routing.
    2.  **Showdown Value Threshold**: The key decision boundary at **Node #21** is `Phys: Kicker`. The Test Case has a Kicker of 0.64 (King), placing it firmly in the **S.D. Value** bucket alongside Kh9h. It is too strong to turn into a pure bluff (like the Th8h) because it beats the villain's air range, but it is too weak to value bet for three streets.
    3.  **Impact Interpretation**: While the Test Case shows positive impact for `Hand EV` and `Kicker`, these summaries in the context of King-high hands on this board promote *Equity Realization* (Checking) over *Equity Denial* (Betting). The hand serves to protect the checking range rather than polarizing the betting range.
\end{promptlisting}

Template 2:
\begin{promptlisting}
# Role
You are an elite Poker AI Strategist and GTO Solver Analyst. Your task is to reverse-engineer the decision-making process exposed by a Mixed-Strategy Decision Tree (MDT) model.

# Task
Analyze the provided trace logic for the specific hand provided below. You must explain *why* the AI chose this specific path over others, using Poker Theory concepts (Equity Realization, Blockers, Range Morphology).

# Semantic Definitions (Crucial)
The trace uses specific tags based on Equity (EQ) vs. Opponent Range:
- **(Value)**: Aggressive action with High EQ (>65%).
- **(Bluff)**: Aggressive action with Low EQ (<35%).
- **(Thin Value/Protect)**: Aggressive action with Moderate EQ.
- **(S.D. Value)**: Passive action (Check/Call) with enough EQ to win at showdown but not enough to bet for value.
- **(Give Up)**: Passive action with near-zero EQ.
- **(Trap)**: Passive action with Nut-class EQ.

# Input Data
=== SCENARIO CONTEXT ===
Scenario : BTN_vs_BB_3B
Board    : As7s6h
History  : k
Player   : IP
Hand     : 8h8c
Hand Wgt : 0.0068 (Prob in Range)
Pot      : 26.5
Actions  : [Check, Fold, Bet(0.48x)]

=== SUMMARY GLOSSARY (Definitions for current path) ===

--- Strategic Adv ---
- **Nut EQ Adv (Hero-Vill)** (Avg: 0.05): Difference in Felting Range Equity. Positive = Hero has stronger nuts.
- **Hand EQ** (Avg: 0.49): Raw Equity (0.0-1.0) against opponent's current range.
- **EQ - MDF** (Avg: -0.20): Hand Equity minus Decision MDF. >0 implies raw equity is sufficient to call.
- **EQ - Range Avg** (Avg: -0.00): Hand EQ minus Range Average EQ. Relative Strength.
- **Rank in Range (EQ)** (Avg: 0.50): Percentile of Hand Equity (0-1).

--- Hand Physics ---
- **Phys: Kicker** (Avg: 0.65): Normalized Kicker Strength (Rank / 14.0). Ace=1.0, King=0.92, ..., 2=0.14. Crucial for domination issues (e.g., distinguishing Top Pair Top Kicker from Top Pair Weak Kicker).

--- Range Buckets ---
- **IP Rng: Set** (Avg: 0.03): Density (0.0-1.0) of Set (ID 12) in IP's range. Sum of all Made Hand densities is approximately 1.0. High value means range is concentrated on this hand type.
- **OOP Rng EV** (Avg: 0.59): OOP's EV expressed as a percentage of the total pot.

--- Other ---
- **Bluff Efficiency** (Avg: 0.61): Measure of how well our 'Air' range blocks opponent's calling range. High Efficiency = We have 'Natural Bluffs' (Low Equity + High Blocker Score). Low Efficiency = Our bluffs have poor removal (random trash).
- **Hand: AceHigh** (Avg: 0.21): Is the current hand a AceHigh? (1.0 = Yes, 0.0 = No).
- **Hand Unblocker** (Avg: 3.62): Score representing how much this hand unblocks opponent's folding range.
- **Opp Card Scarcity (Avg)** (Avg: 0.06): Average Scarcity Effect of our hole cards. Measures the 'Scarcity Effect' we inflict on the opponent by holding cards they need. Calculated based on card frequency in their range. High Value = The opponent's range heavily relies on this card. By holding it, we create a severe shortage (High Blocker Power). Low Value = The opponent's range rarely contains this card. Holding it creates minimal shortage (Low Blocker Power).
- **Opp Card Scarcity (C1)** (Avg: 0.07): Scarcity Effect of Card 1. High Value = We block a key card for the opponent (e.g. holding an Ace vs an Ace-heavy range).
- **Hand EV** (Avg: 0.41): Expected Value of the hand normalized by Pot.
- **Hand Vuln** (Avg: 0.02): Probability (0-1) that hand is currently ahead but will lose by river.
- **MDF (Decision)** (Avg: 0.65): Decision-based MDF = Pot / (Pot + Bet). The break-even equity required to call.
- **percentage of hand in own range** (Avg: 0.01): Percentage of hand in own range
- **Regret: Call-Raise** (Avg: 0.34): EV(Call) - EV(Raise). Positive = Call is better. Diff in EV (Action A - Action B) normalized by Pot.

=== DECISION LOGIC TRACE ===
Legend:
 - [Path]: format is 'Role [Hand] -> PathID -> IntendedAction'. Indicates which internal branch was taken and the final action intent.
 - (Reach): The probability of the hand actually reaching this node vs. surviving to the next node.
 - R: [0:xx 1:xx ...]: Router Probabilities. The internal neural network's confidence distribution across Branch 0, 1, and 2. Shows how 'split' or 'certain' the decision was.
 - [Summary Impact]: Shows which summaries pushed the router towards specific branches.
+-- NODE #0
      [Path] Hero    [8h8c] -> P2->Check  (Reach: 100%->52%) | R: [0:48 1:00 2:51]
      [Path]  Shadow1 [JsTs] -> P2->Bet    (Reach: 100%->80%) | R: [0:20 1:00 2:79]
      [Path]  Shadow2 [KsQh] -> P0->Check  (Reach: 100%->83%) | R: [0:82 1:00 2:17]
      [Path]  Shadow3 [QsJs] -> P2->Bet    (Reach: 100%->84%) | R: [0:16 1:00 2:83]
      [Summary Impact Analysis - Node #0]
      | Summary                   | Avg      | Hero Impact (Bet-Chk)               | Shadow1      | Shadow2      | Shadow3      |
      | :------------------------ | :------: | :---------------------------------: | :----------: | :----------: | :----------: |
      | MDF (Decision)            | 0.65     | **-1.00** (+6.3)                    | -1.00        | -1.00        | -1.00        |
      | Rank in Range (EQ)        | 0.50     | **0.43** (-6.0)                     | 0.55         | 0.38         | 0.59         |
      | Phys: Kicker              | 0.65     | **0.57** (+2.1)                     | 0.71         | 0.86         | 0.79         |
      | Hand EQ                   | 0.49     | **0.41** (-1.6)                     | 0.48         | 0.40         | 0.52         |
      | Hand EV                   | 0.41     | **0.29** (-0.2)                     | 0.71         | 0.29         | 0.75         |
    +-- NODE #1
    |     [Path]  Shadow2 [KsQh] -> P2->Check  (Reach: 83%->100%) | R: [0:00 1:00 2:99]
    |     [Summary Impact Analysis - Node #1 (Analysis Focus: Shadow2 - Hero diverged)]
    |     | Summary                   | Avg      | Shadow2 Impact (Bet-(Chk))          |
    |     | :------------------------ | :------: | :---------------------------------: |
    |     | Hand Blocker              | 3.62     | **4.00** (+37.9)                    |
    |     | Hero EV Adv               | -0.19    | **0.17** (-33.3)                    |
    |     | Hand EV                   | 0.41     | **0.29** (+26.8)                    |
    |     | Vuln - Range Avg          | 0.00     | **-0.06** (+13.8)                   |
    |     | Hand EQ                   | 0.49     | **0.40** (+3.3)                     |
    |   +-- NODE #6
    |         [Path]  Shadow2 [KsQh] -> P2->Check  (Reach: 83%->100%) | R: [0:00 1:00 2:99]
    |         [Summary Impact Analysis - Node #6 (Analysis Focus: Shadow2 - Hero diverged)]
    |         | Summary                   | Avg      | Shadow2 Impact (Bet-(Chk))          |
    |         | :------------------------ | :------: | :---------------------------------: |
    |         | Hand EV                   | 0.41     | **0.29** (+46.8)                    |
    |         | Hero EQ Adv               | -0.03    | **0.04** (-12.5)                    |
    |         | Phys: Kicker              | 0.65     | **0.86** (+7.0)                     |
    |         | Bluff Efficiency          | 0.61     | **0.56** (+6.8)                     |
    |         | EQ - Nut Range Avg        | 0.27     | **0.16** (+5.9)                     |
    |       +-- NODE #21
    |             [Path]  Shadow2 [KsQh] -> P0->Check  (Reach: 83%->99%) | R: [0:99 1:00 2:00]
    |             [Summary Impact Analysis - Node #21 (Analysis Focus: Shadow2 - Hero diverged)]
    |             | Summary                   | Avg      | Shadow2 Impact (Chk-Bet)            |
    |             | :------------------------ | :------: | :---------------------------------: |
    |             | Phys: Kicker              | 0.65     | **0.86** (+17.9)                    |
    |             | Vuln - Range Avg          | 0.00     | **-0.06** (-1.2)                    |
    |             | EQ - Nut MDF              | 0.25     | **0.17** (-0.7)                     |
    |             | OOP Rng EV                | 0.59     | **0.42** (+0.6)                     |
    |             | Regret: Fold-Call         | 0.89     | **1.16** (+0.4)                     |
    |           +-- LEAF #64 (Static Prototype)
    |                 -> Strat: [Check:99%] | Samples: [Shadow2:KsQh] (Reach:82.3%)
    +-- NODE #3
          [Path] Hero    [8h8c] -> P0->Check  (Reach: 52%->79%) | R: [0:78 1:00 2:21]
          [Path]  Shadow1 [JsTs] -> P0->Bet    (Reach: 80%->80%) | R: [0:79 1:00 2:20]
          [Path]  Shadow3 [QsJs] -> P0->Bet    (Reach: 84%->70%) | R: [0:70 1:00 2:29]
          [Summary Impact Analysis - Node #3]
          | Summary                   | Avg      | Hero Impact (Chk-Bet)               | Shadow1      | Shadow3      |
          | :------------------------ | :------: | :---------------------------------: | :----------: | :----------: |
          | Regret: Call-Raise        | 0.34     | **0.33** (+6.1)                     | 0.33         | 0.27         |
          | EQ - MDF                  | -0.20    | **-1.00** (-2.0)                    | -1.00        | -1.00        |
          | IP Rng: Set               | 0.03     | **0.05** (+1.2)                     | 0.05         | 0.05         |
          | Hand: AceHigh             | 0.21     | **0.00** (-0.6)                     | 0.00         | 0.00         |
          | Bluff Efficiency          | 0.61     | **0.56** (+0.3)                     | 0.56         | 0.56         |
        +-- NODE #10
              [Path] Hero    [8h8c] -> P2->Check  (Reach: 41%->100%) | R: [0:00 1:00 2:99]
              [Path]  Shadow1 [JsTs] -> P0->Bet    (Reach: 63%->100%) | R: [0:99 1:00 2:00]
              [Path]  Shadow3 [QsJs] -> P0->Bet    (Reach: 59%->100%) | R: [0:99 1:00 2:00]
              [Summary Impact Analysis - Node #10]
              | Summary                   | Avg      | Hero Impact (Bet-(Chk))             | Shadow1      | Shadow3      |
              | :------------------------ | :------: | :---------------------------------: | :----------: | :----------: |
              | EQ - Range Avg            | -0.00    | **-0.11** (+14.0)                   | -0.04        | 0.00         |
              | Hand Vuln                 | 0.02     | **0.18** (+5.0)                     | -0.36        | -0.40        |
              | Nut EQ Adv (Hero-Vill)    | 0.05     | **-0.04** (-1.2)                    | -0.04        | -0.04        |
              | EQ - MDF                  | -0.20    | **-1.00** (-0.4)                    | -1.00        | -1.00        |
              | Hand Unblocker            | 3.62     | **1.00** (-0.1)                     | 3.00         | 3.00         |
            +-- NODE #31
            |     [Path]  Shadow1 [JsTs] -> P2->Bet    (Reach: 63%->90%) | R: [0:09 1:00 2:89]
            |     [Path]  Shadow3 [QsJs] -> P2->Bet    (Reach: 59%->98%) | R: [0:02 1:00 2:97]
            |     [Summary Impact Analysis - Node #31 (Analysis Focus: Shadow1 - Hero diverged)]
            |     | Summary                   | Avg      | Shadow1 Impact (Bet-Chk)            | Shadow3      |
            |     | :------------------------ | :------: | :---------------------------------: | :----------: |
            |     | Regret: Fold-Call         | 0.89     | **0.47** (+8.4)                     | 0.40         |
            |     | Block Opp EQ              | 0.51     | **0.50** (-7.8)                     | 0.53         |
            |     | Bet Morph (EQ Diff)       | 0.30     | **0.43** (+1.6)                     | 0.43         |
            |     | Nut EQ Adv (Hero-Vill)    | 0.05     | **-0.04** (+1.5)                    | -0.04        |
            |     | Hand: NoDraw              | 0.52     | **0.00** (-0.4)                     | 0.00         |
            |   +-- LEAF #96 (Static Prototype)
            |         -> Strat: [Bet:99%] | Samples: [Shadow1:JsTs] (Reach:56.8%), [Shadow3:QsJs] (Reach:57.8%)
            +-- NODE #33
                  [Path] Hero    [8h8c] -> P2->Check  (Reach: 41%->99%) | R: [0:00 1:00 2:99]
                  [Summary Impact Analysis - Node #33]
                  | Summary                   | Avg      | Hero Impact (Bet-Chk)               |
                  | :------------------------ | :------: | :---------------------------------: |
                  | Opp Card Scarcity (Avg)   | 0.06     | **0.04** (+12.1)                    |
                  | Opp Card Scarcity (C1)    | 0.07     | **0.04** (-2.1)                     |
                  | MDF (Decision)            | 0.65     | **-1.00** (-1.8)                    |
                  | OOP Rng EV                | 0.59     | **0.42** (-0.9)                     |
                  | percentage of hand in own range | 0.01     | **0.01** (+0.4)                     |
                +-- LEAF #102 (Static Prototype)
                      -> Strat: [Bet:99%] | Samples: [Hero:8h8c] (Reach:40.7%)

=== FINAL STRATEGY vs GTO COMPARISON (Training Data) ===
| Hand     | Role     | Final Model Strat    | GTO Target Strat     |
| :------: | :------: | :------------------: | :------------------: |
| 8h8c     | Hero     | C:0.65 F:0.00 B:0.35 | C:0.63 F:0.00 B:0.37 |
| JsTs     | Shadow1  | C:0.48 F:0.00 B:0.52 | C:0.45 F:0.00 B:0.55 |
| KsQh     | Shadow2  | C:0.60 F:0.00 B:0.40 | C:0.79 F:0.00 B:0.21 |
| QsJs     | Shadow3  | C:0.46 F:0.00 B:0.54 | C:0.47 F:0.00 B:0.53 |

## 3. Test Case: The Unseen Hand
Consider a new hand in the same scenario:
- **Hand**: 8s8c
- **Key Difference**: (Inspect the summaries yourself compared to Hero)
**Question**: Based on the logic learned above, what is the optimal action for this hand? Explain why using the model's decision boundaries.
# Analysis Instructions
1. **Layer-by-Layer Review**: For each NODE, explain how the specific **Impact values** determined the branch choice.
2. **Impact Comparison**: Compare Hero's summary values vs. Global Avg and Shadows.
3. **Consistency Check**: Use Section 3 (Test Case) to verify if the logic you reverse-engineered applies to an unseen hand.
4. **Solve Test Case**: Provide your answer for the Unseen Hand in Section 3.
    - Key Summary Values of Test Case (Hierarchical):
    --- **(Depth 0)** - Top 5 Drivers:
      - **MDF (Decision)**: -1.00 (Avg: 0.65) | Impact: +6.30
      - **Rank in Range (EQ)**: 0.45 (Avg: 0.50) | Impact: -4.42
      - **Phys: Kicker**: 0.57 (Avg: 0.65) | Impact: +2.13
      - **Hand EV**: 0.33 (Avg: 0.41) | Impact: +0.35
      - **Hand EQ**: 0.44 (Avg: 0.49) | Impact: +0.09
    --- **(Depth 1)** - Top 4 Drivers:
      - **Regret: Call-Raise**: 0.39 (Avg: 0.34) | Impact: +8.67
      - **EQ - MDF**: -1.00 (Avg: -0.20) | Impact: -2.04
      - **IP Rng: Set**: 0.05 (Avg: 0.03) | Impact: +1.19
      - **Hand: AceHigh**: 0.00 (Avg: 0.21) | Impact: -0.61
    --- **(Depth 2)** - Top 3 Drivers:
      - **EQ - Range Avg**: -0.08 (Avg: -0.00) | Impact: +12.29
      - **Hand Vuln**: 0.16 (Avg: 0.02) | Impact: +2.25
      - **Nut EQ Adv (Hero-Vill)**: -0.04 (Avg: 0.05) | Impact: -1.16
    --- **(Depth 3)** - Top 2 Drivers:
      - **Opp Card Scarcity (Avg)**: 0.03 (Avg: 0.06) | Impact: +18.36
      - **MDF (Decision)**: -1.00 (Avg: 0.65) | Impact: -1.85

# Output Format (Markdown)
## 1. Strategy Summary
(One concise sentence)

## 2. Path Logic & Divergence
- **Node #0**: ...

## 3. Test Case Solution
- **Predicted Mixed Strategy**: ...
- **Reasoning**: ...
\end{promptlisting}

Answer 2:
\begin{promptlisting}
# Test Case Solution (Sample 220974)
Hand: 8s8c
Scenario: BTN_vs_BB_3B

## 1. Ground Truth Strategies
- **GTO Strategy**:   C:0.56 F:0.00 B:0.44
- **Model Strategy**: C:0.63 F:0.00 B:0.37
## 2. Actual Decision Trace
+-- NODE #0
      [Path]  TestHand [8s8c] -> P2->Check  (Reach: 100%->66%) | R: [0:34 1:00 2:65]
      [Summary Impact Analysis - Node #0 (Analysis Focus: TestHand - Hero diverged)]
      | Summary                   | Avg      | TestHand Impact (Bet-Chk)           |
      | :------------------------ | :------: | :---------------------------------: |
      | MDF (Decision)            | 0.65     | **-1.00** (+6.3)                    |
      | Rank in Range (EQ)        | 0.50     | **0.45** (-4.4)                     |
      | Phys: Kicker              | 0.65     | **0.57** (+2.1)                     |
      | Hand EV                   | 0.41     | **0.33** (+0.4)                     |
      | Hand EQ                   | 0.49     | **0.44** (+0.1)                     |
    +-- NODE #3
          [Path]  TestHand [8s8c] -> P0->Check  (Reach: 66%->87%) | R: [0:87 1:00 2:12]
          [Summary Impact Analysis - Node #3 (Analysis Focus: TestHand - Hero diverged)]
          | Summary                   | Avg      | TestHand Impact (Chk-Bet)           |
          | :------------------------ | :------: | :---------------------------------: |
          | Regret: Call-Raise        | 0.34     | **0.39** (+8.7)                     |
          | EQ - MDF                  | -0.20    | **-1.00** (-2.0)                    |
          | IP Rng: Set               | 0.03     | **0.05** (+1.2)                     |
          | Hand: AceHigh             | 0.21     | **0.00** (-0.6)                     |
          | Bluff Efficiency          | 0.61     | **0.56** (+0.3)                     |
        +-- NODE #10
              [Path]  TestHand [8s8c] -> P2->Check  (Reach: 57%->100%) | R: [0:00 1:00 2:99]
              [Summary Impact Analysis - Node #10 (Analysis Focus: TestHand - Hero diverged)]
              | Summary                   | Avg      | TestHand Impact (Bet-(Chk))         |
              | :------------------------ | :------: | :---------------------------------: |
              | EQ - Range Avg            | -0.00    | **-0.08** (+12.3)                   |
              | Hand Vuln                 | 0.02     | **0.16** (+2.2)                     |
              | Nut EQ Adv (Hero-Vill)    | 0.05     | **-0.04** (-1.2)                    |
              | EQ - MDF                  | -0.20    | **-1.00** (-0.4)                    |
              | Hand Unblocker            | 3.62     | **1.00** (-0.1)                     |
            +-- NODE #33
                  [Path]  TestHand [8s8c] -> P2->Check  (Reach: 57%->100%) | R: [0:00 1:00 2:99]
                  [Summary Impact Analysis - Node #33 (Analysis Focus: TestHand - Hero diverged)]
                  | Summary                   | Avg      | TestHand Impact (Bet-(Chk))         |
                  | :------------------------ | :------: | :---------------------------------: |
                  | Opp Card Scarcity (Avg)   | 0.06     | **0.03** (+18.4)                    |
                  | MDF (Decision)            | 0.65     | **-1.00** (-1.8)                    |
                  | Opp Card Scarcity (C1)    | 0.07     | **0.03** (-1.2)                     |
                  | OOP Rng EV                | 0.59     | **0.42** (-0.9)                     |
                  | percentage of hand in own range | 0.01     | **0.01** (+0.4)                     |
                +-- LEAF #102 (Static Prototype)
                      -> Strat: [Bet:99%] | Samples: [TestHand:8s8c] (Reach:57.4%)

## 3. Key Summary Drivers (Path-Dependent)

### **(Depth 0)** - Top 5 Drivers:
- **MDF (Decision)**: -1.00 (Avg: 0.65) | Impact: +6.30
- **Rank in Range (EQ)**: 0.45 (Avg: 0.50) | Impact: -4.42
- **Phys: Kicker**: 0.57 (Avg: 0.65) | Impact: +2.13
- **Hand EV**: 0.33 (Avg: 0.41) | Impact: +0.35
- **Hand EQ**: 0.44 (Avg: 0.49) | Impact: +0.09

### **(Depth 1)** - Top 4 Drivers:
- **Regret: Call-Raise**: 0.39 (Avg: 0.34) | Impact: +8.67
- **EQ - MDF**: -1.00 (Avg: -0.20) | Impact: -2.04
- **IP Rng: Set**: 0.05 (Avg: 0.03) | Impact: +1.19
- **Hand: AceHigh**: 0.00 (Avg: 0.21) | Impact: -0.61

### **(Depth 2)** - Top 3 Drivers:
- **EQ - Range Avg**: -0.08 (Avg: -0.00) | Impact: +12.29
- **Hand Vuln**: 0.16 (Avg: 0.02) | Impact: +2.25
- **Nut EQ Adv (Hero-Vill)**: -0.04 (Avg: 0.05) | Impact: -1.16

### **(Depth 3)** - Top 2 Drivers:
- **Opp Card Scarcity (Avg)**: 0.03 (Avg: 0.06) | Impact: +18.36
- **MDF (Decision)**: -1.00 (Avg: 0.65) | Impact: -1.85
\end{promptlisting}

LLM Response 2:
\begin{promptlisting}
## 1. Strategy Summary

The AI plays  as a **mixed strategy** (Check ~65% / Bet ~35%), primarily checking to realize equity with marginal showdown value, but betting occasionally to deny equity (Protection) from overcards when blocker configurations allow.

---

## 2. Path Logic & Divergence

### **Node #0: The Strategic Split (Protection vs. Pot Control)**

* **Logic**: The AI is torn between two competing incentives.
* **Pull to Check (Path P0)**: Driven by **Rank in Range (0.43)** and **Hand EQ (0.49)**.  is a mediocre hand ( percentile) on an  board. It cannot withstand heavy heat (indicated by negative MDF impact), favoring a low-variance line.
* **Pull to Bet (Path P2)**: Driven by **Phys: Kicker (+2.1 Impact)** and **MDF Decision (+6.3 Impact)**. Despite being marginal,  is highly **Vulnerable** to overcards. The positive MDF impact here suggests that checking gives up too much equity to random live cards, incentivizing a "merging" bet to fold out overcards (KQ, KJ, QJ).


* **Divergence**: Shadow hands like  (Draws) commit fully to Path P2 (Bet) because they have high **Hand EV** and **Bluff Efficiency**. Hero () lacks the "Natural Bluff" properties, causing the split.

### **Node #33: The Blocker Confirmation (Why Bet?)**

* **Path**: Inside the "Bet" branch (P2), Hero reaches Node #33.
* **Logic**: The decision to finalize the Bet is driven by **Opp Card Scarcity (Avg)** (+12.1 Impact) and **Hand Vuln**.
* The AI recognizes that  is incredibly vulnerable (Impact: +5.0 in previous Node #10).
* The **Scarcity** metric indicates that  blocks key portions of the opponent's range (likely straights like 98, 85, or T987 combos). By blocking these static hands, Hero increases the likelihood the opponent holds "Air" or dominatable pairs, making the **Protection Bet** profitable.



### **Node #3: The Showdown Realization (Why Check?)**

* **Path**: Inside the "Check" branch (P0).
* **Logic**: Here, **Regret: Call-Raise (+6.1 Impact)** dominates. The AI calculates that if it Bets and gets Raised, it faces a catastrophic EV loss. By Checking, it avoids this "Regret" scenario, successfully realizing its equity against the opponent's garbage range without paying to see the turn.

---

## 3. Test Case Solution

**Test Hand**:  (Holding the Spade Blocker)
**Context**: Same Board ()

### **Predicted Mixed Strategy**

**Bet: 55% / Check: 45%** (Significant increase in Betting frequency compared to )

### **Reasoning**

The decision hinges on the **"Opp Card Scarcity"** and **"Hand Unblocker"** mechanics acting on the Flush Draw board ().

1. **Increased Fold Equity (Blocker Effect)**:
* In the provided Test Case data, **Opp Card Scarcity** has a massive impact of **+18.36** towards Betting (Depth 3), significantly higher than the +12.1 seen for .
* **Why?** Holding the  blocks the opponent's "Continuing Range" (Flush Draws like ).
* By blocking the hands that naturally call, the opponent's remaining range becomes more polarized between "Air" (which folds) and "Ax" (which crushes us).


2. **Protection Incentive**:
* Since  is a vulnerable hand, it benefits immensely from immediate folds (denying equity to random overcards like ).
* With the , the opponent folds more frequently (because we block their calling draws). This makes the **Protection Bet** more efficient than with .


3. **Conclusion**: The logic shifts from "Check to avoid draws" (Red) to "Bet to capitalize on fold equity" (Blue). The  transforms the hand from a passive showdown hand into a more effective semi-bluff/protection candidate.
\end{promptlisting}

Template 3:
\begin{promptlisting}
# Role
You are an elite Poker AI Strategist and GTO Solver Analyst. Your task is to reverse-engineer the decision-making process exposed by a Mixed-Strategy Decision Tree (MDT) model.

# Task
Analyze the provided trace logic for the specific hand provided below. You must explain *why* the AI chose this specific path over others, using Poker Theory concepts (Equity Realization, Blockers, Range Morphology).

# Semantic Definitions (Crucial)
The trace uses specific tags based on Equity (EQ) vs. Opponent Range:
- **(Value)**: Aggressive action with High EQ (>65%).
- **(Bluff)**: Aggressive action with Low EQ (<35%).
- **(Thin Value/Protect)**: Aggressive action with Moderate EQ.
- **(S.D. Value)**: Passive action (Check/Call) with enough EQ to win at showdown but not enough to bet for value.
- **(Give Up)**: Passive action with near-zero EQ.
- **(Trap)**: Passive action with Nut-class EQ.

# Input Data
=== SCENARIO CONTEXT ===
Scenario : BTN_vs_BB_3B
Board    : QsJh5d
History  : k
Player   : IP
Hand     : 7c6c
Hand Wgt : 0.0048 (Prob in Range)
Pot      : 26.5
Actions  : [Check, Fold, Bet(0.67x)]

=== SUMMARY GLOSSARY (Definitions for current path) ===

--- Strategic Adv ---
- **Hero EV Adv** (Avg: -0.19): **RELATIVE SUMMARY (Hero - Villain)**. Difference in EV Pot Share. Positive = Favors Hero.
- **Hand EQ** (Avg: 0.49): Raw Equity (0.0-1.0) against opponent's current range.
- **EQ if BetBet** (Avg: 0.35): Hypothetical Equity if the game line goes BetBet.
- **EQ - MDF** (Avg: -0.20): Hand Equity minus Decision MDF. >0 implies raw equity is sufficient to call.
- **Rank in Range (EQ)** (Avg: 0.50): Percentile of Hand Equity (0-1).

--- Hand Physics ---
- **Phys: Kicker** (Avg: 0.65): Normalized Kicker Strength (Rank / 14.0). Ace=1.0, King=0.92, ..., 2=0.14. Crucial for domination issues (e.g., distinguishing Top Pair Top Kicker from Top Pair Weak Kicker).

--- Other ---
- **Hand: NoDraw** (Avg: 0.52): Does the hand have a NoDraw? (1.0 = Yes, 0.0 = No).
- **Hand Blocker** (Avg: 3.62): Score representing how much this hand blocks opponent's continuing range.
- **Block Self Realiz** (Avg: 0.49): Self-Blocker: Negative effect where our cards block opponent's folding range (bad for bluffs).
- **Opp Card Scarcity (C1)** (Avg: 0.07): Scarcity Effect of Card 1. High Value = We block a key card for the opponent (e.g. holding an Ace vs an Ace-heavy range).
- **Hand EV** (Avg: 0.41): Expected Value of the hand normalized by Pot.
- **MDF (Decision)** (Avg: 0.65): Decision-based MDF = Pot / (Pot + Bet). The break-even equity required to call.
- **Blocker - Range Avg** (Avg: 0.00): Hand Blocker Score minus Range Average. Positive = Better than average blockers.
- **Vuln - Range Avg** (Avg: 0.00): Hand Vulnerability minus Range Average. Positive = More vulnerable than average (needs protection).
- **percentage of hand in own range** (Avg: 0.01): Percentage of hand in own range
- **Regret: Fold-Call** (Avg: 0.89): EV(Fold) - EV(Call). Positive = Fold is better. Diff in EV (Action A - Action B) normalized by Pot.

=== DECISION LOGIC TRACE ===
Legend:
 - [Path]: format is 'Role [Hand] -> PathID -> IntendedAction'. Indicates which internal branch was taken and the final action intent.
 - (Reach): The probability of the hand actually reaching this node vs. surviving to the next node.
 - R: [0:xx 1:xx ...]: Router Probabilities. The internal neural network's confidence distribution across Branch 0, 1, and 2. Shows how 'split' or 'certain' the decision was.
 - [Summary Impact]: Shows which summaries pushed the router towards specific branches.
+-- NODE #0
      [Path] Hero    [7c6c] -> P0->Check  (Reach: 100%->99%) | R: [0:98 1:00 2:01]
      [Path]  Shadow1 [8s7s] -> P0->Check  (Reach: 100%->95%) | R: [0:95 1:00 2:04]
      [Path]  Shadow2 [7s6s] -> P0->Bet    (Reach: 100%->96%) | R: [0:95 1:00 2:04]
      [Path]  Shadow3 [8h7h] -> P0->Bet    (Reach: 100%->95%) | R: [0:94 1:00 2:05]
      [Summary Impact Analysis - Node #0]
      | Summary                   | Avg      | Hero Impact (Chk-Bet)               | Shadow1      | Shadow2      | Shadow3      |
      | :------------------------ | :------: | :---------------------------------: | :----------: | :----------: | :----------: |
      | Rank in Range (EQ)        | 0.50     | **0.00** (+31.9)                    | 0.03         | 0.01         | 0.04         |
      | Hand EQ                   | 0.49     | **0.16** (+25.4)                    | 0.24         | 0.20         | 0.24         |
      | MDF (Decision)            | 0.65     | **-1.00** (-6.3)                    | -1.00        | -1.00        | -1.00        |
      | Phys: Kicker              | 0.65     | **0.43** (-4.1)                     | 0.50         | 0.43         | 0.50         |
      | Hand EV                   | 0.41     | **0.29** (+0.4)                     | 0.34         | 0.35         | 0.34         |
    +-- NODE #1
          [Path] Hero    [7c6c] -> P0->Check  (Reach: 99%->54%) | R: [0:54 1:00 2:45]
          [Path]  Shadow1 [8s7s] -> P2->Check  (Reach: 95%->52%) | R: [0:47 1:00 2:52]
          [Path]  Shadow2 [7s6s] -> P2->Bet    (Reach: 96%->52%) | R: [0:47 1:00 2:52]
          [Path]  Shadow3 [8h7h] -> P2->Bet    (Reach: 95%->52%) | R: [0:47 1:00 2:52]
          [Summary Impact Analysis - Node #1]
          | Summary                   | Avg      | Hero Impact (Chk-Bet)               | Shadow1      | Shadow2      | Shadow3      |
          | :------------------------ | :------: | :---------------------------------: | :----------: | :----------: | :----------: |
          | Hero EV Adv               | -0.19    | **0.27** (+44.9)                    | 0.27         | 0.27         | 0.27         |
          | Hand EV                   | 0.41     | **0.29** (-25.8)                    | 0.34         | 0.35         | 0.34         |
          | Vuln - Range Avg          | 0.00     | **-0.08** (-14.4)                   | -0.15        | -0.11        | -0.15        |
          | Hand EQ                   | 0.49     | **0.16** (+1.9)                     | 0.24         | 0.20         | 0.24         |
          | Hand Blocker              | 3.62     | **1.00** (-1.7)                     | 1.00         | 1.00         | 1.00         |
        +-- NODE #4
        |     [Path] Hero    [7c6c] -> P2->Check  (Reach: 54%->100%) | R: [0:00 1:00 2:99]
        |     [Summary Impact Analysis - Node #4]
        |     | Summary                   | Avg      | Hero Impact (Bet-(Chk))             |
        |     | :------------------------ | :------: | :---------------------------------: |
        |     | percentage of hand in own range | 0.01     | **0.00** (+68.5)                    |
        |     | Block Self Realiz         | 0.49     | **0.39** (-4.7)                     |
        |     | Blocker - Range Avg       | 0.00     | **-0.27** (+2.0)                    |
        |     | EQ if BetBet              | 0.35     | **0.12** (-1.1)                     |
        |     | EQ - MDF                  | -0.20    | **-1.00** (-0.6)                    |
        |   +-- NODE #15
        |         [Path] Hero    [7c6c] -> P0->Check  (Reach: 54%->100%) | R: [0:99 1:00 2:00]
        |         [Summary Impact Analysis - Node #15]
        |         | Summary                   | Avg      | Hero Impact (Chk-Fld)               |
        |         | :------------------------ | :------: | :---------------------------------: |
        |         | Hero EV Adv               | -0.19    | **0.27** (-22.6)                    |
        |         | Hand: NoDraw              | 0.52     | **0.00** (+15.7)                    |
        |         | EQ if BetBet              | 0.35     | **0.12** (+12.0)                    |
        |         | Opp Card Scarcity (C1)    | 0.07     | **0.04** (-9.1)                     |
        |         | Regret: Fold-Call         | 0.89     | **1.51** (+1.0)                     |
        |       +-- LEAF #46 (Static Prototype)
        |             -> Strat: [Check:13%/Bet:86%] | Samples: [Hero:7c6c] (Reach:53.6%)
        +-- NODE #6
              [Path]  Shadow1 [8s7s] -> P2->Check  (Reach: 50%->42%) | R: [0:28 1:28 2:42]
              [Path]  Shadow2 [7s6s] -> P2->Bet    (Reach: 50%->54%) | R: [0:23 1:23 2:53]
              [Path]  Shadow3 [8h7h] -> P2->Bet    (Reach: 50%->43%) | R: [0:28 1:28 2:42]
              [Summary Impact Analysis - Node #6 (Analysis Focus: Shadow1 - Hero diverged)]
              | Summary                   | Avg      | Shadow1 Impact (Bet-Chk)            | Shadow2      | Shadow3      |
              | :------------------------ | :------: | :---------------------------------: | :----------: | :----------: |
              | Hand EV                   | 0.41     | **0.34** (+54.6)                    | 0.35         | 0.34         |
              | Hero EQ Adv               | -0.03    | **0.17** (-35.6)                    | 0.17         | 0.17         |
              | Phys: Kicker              | 0.65     | **0.50** (+7.9)                     | 0.43         | 0.50         |
              | Bluff Efficiency          | 0.61     | **0.63** (+6.8)                     | 0.63         | 0.63         |
              | EQ - Nut Range Avg        | 0.27     | **-0.02** (+4.4)                    | -0.06        | -0.01        |
            +-- NODE #21
                  [Path]  Shadow1 [8s7s] -> P2->Check  (Reach: 21%->97%) | R: [0:03 1:00 2:96]
                  [Path]  Shadow2 [7s6s] -> P2->Bet    (Reach: 27%->98%) | R: [0:01 1:00 2:98]
                  [Path]  Shadow3 [8h7h] -> P2->Bet    (Reach: 21%->97%) | R: [0:02 1:00 2:97]
                  [Summary Impact Analysis - Node #21 (Analysis Focus: Shadow1 - Hero diverged)]
                  | Summary                   | Avg      | Shadow1 Impact (Bet-Chk)            | Shadow2      | Shadow3      |
                  | :------------------------ | :------: | :---------------------------------: | :----------: | :----------: |
                  | EQ - Nut MDF              | 0.25     | **0.00** (+6.5)                     | -0.04        | 0.01         |
                  | Regret: Fold-Call         | 0.89     | **1.43** (-6.5)                     | 1.42         | 1.43         |
                  | Phys: Kicker              | 0.65     | **0.50** (+3.8)                     | 0.43         | 0.50         |
                  | Vuln - Range Avg          | 0.00     | **-0.15** (+3.6)                    | -0.11        | -0.15        |
                  | OOP Rng EV                | 0.59     | **0.36** (-2.3)                     | 0.36         | 0.36         |
                +-- LEAF #66 (Static Prototype)
                      -> Strat: [Check:99%] | Samples: [Shadow1:8s7s] (Reach:20.2%), [Shadow2:7s6s] (Reach:26.3%), [Shadow3:8h7h] (Reach:20.7%)

=== FINAL STRATEGY vs GTO COMPARISON (Training Data) ===
| Hand     | Role     | Final Model Strat    | GTO Target Strat     |
| :------: | :------: | :------------------: | :------------------: |
| 7c6c     | Hero     | C:0.71 F:0.00 B:0.29 | C:0.71 F:0.00 B:0.29 |
| 8s7s     | Shadow1  | C:0.56 F:0.00 B:0.44 | C:0.59 F:0.00 B:0.41 |
| 7s6s     | Shadow2  | C:0.42 F:0.00 B:0.58 | C:0.43 F:0.00 B:0.57 |
| 8h7h     | Shadow3  | C:0.48 F:0.00 B:0.52 | C:0.42 F:0.00 B:0.58 |

## 3. Test Case: The Unseen Hand
Consider a new hand in the same scenario:
- **Hand**: 8d7d
- **Key Difference**: (Inspect the summaries yourself compared to Hero)
**Question**: Based on the logic learned above, what is the optimal action for this hand? Explain why using the model's decision boundaries.
# Analysis Instructions
1. **Layer-by-Layer Review**: For each NODE, explain how the specific **Impact values** determined the branch choice.
2. **Impact Comparison**: Compare Hero's summary values vs. Global Avg and Shadows.
3. **Consistency Check**: Use Section 3 (Test Case) to verify if the logic you reverse-engineered applies to an unseen hand.
4. **Solve Test Case**: Provide your answer for the Unseen Hand in Section 3.
    - Key Summary Values of Test Case (Hierarchical):
    --- **(Depth 0)** - Top 5 Drivers:
      - **Rank in Range (EQ)**: 0.03 (Avg: 0.50) | Impact: +30.35
      - **Hand EQ**: 0.24 (Avg: 0.49) | Impact: +17.70
      - **MDF (Decision)**: -1.00 (Avg: 0.65) | Impact: -6.30
      - **Phys: Kicker**: 0.50 (Avg: 0.65) | Impact: -3.15
      - **Hand EV**: 0.34 (Avg: 0.41) | Impact: -0.40
    --- **(Depth 1)** - Top 4 Drivers:
      - **Hero EV Adv**: 0.27 (Avg: -0.19) | Impact: -44.88
      - **Hand EV**: 0.34 (Avg: 0.41) | Impact: +30.94
      - **Vuln - Range Avg**: -0.15 (Avg: 0.00) | Impact: +15.60
      - **Hand Blocker**: 1.00 (Avg: 3.62) | Impact: +1.67
    --- **(Depth 2)** - Top 3 Drivers:
      - **Hand EV**: 0.34 (Avg: 0.41) | Impact: +54.60
      - **Hero EQ Adv**: 0.17 (Avg: -0.03) | Impact: -35.60
      - **Phys: Kicker**: 0.50 (Avg: 0.65) | Impact: +7.90
    --- **(Depth 3)** - Top 2 Drivers:
      - **EQ - Nut MDF**: 0.00 (Avg: 0.25) | Impact: +6.50
      - **Regret: Fold-Call**: 1.43 (Avg: 0.89) | Impact: -6.47

# Output Format (Markdown)
## 1. Strategy Summary
(One concise sentence)

## 2. Path Logic & Divergence
- **Node #0**: ...

## 3. Test Case Solution
- **Predicted Mixed Strategy**: ...
- **Reasoning**: ...

\end{promptlisting}

Answer 3:
\begin{promptlisting}
# Test Case Solution (Sample 602381)
Hand: 8d7d
Scenario: BTN_vs_BB_3B

## 1. Ground Truth Strategies
- **GTO Strategy**:   C:0.56 F:0.00 B:0.44
- **Model Strategy**: C:0.56 F:0.00 B:0.44
## 2. Actual Decision Trace
+-- NODE #0
      [Path]  TestHand [8d7d] -> P0->Check  (Reach: 100%->95%) | R: [0:95 1:00 2:04]
      [Summary Impact Analysis - Node #0 (Analysis Focus: TestHand - Hero diverged)]
      | Summary                   | Avg      | TestHand Impact (Chk-Bet)           |
      | :------------------------ | :------: | :---------------------------------: |
      | Rank in Range (EQ)        | 0.50     | **0.03** (+30.4)                    |
      | Hand EQ                   | 0.49     | **0.24** (+17.7)                    |
      | MDF (Decision)            | 0.65     | **-1.00** (-6.3)                    |
      | Phys: Kicker              | 0.65     | **0.50** (-3.1)                     |
      | Hand EV                   | 0.41     | **0.34** (-0.4)                     |
    +-- NODE #1
          [Path]  TestHand [8d7d] -> P2->Check  (Reach: 95%->52%) | R: [0:47 1:00 2:52]
          [Summary Impact Analysis - Node #1 (Analysis Focus: TestHand - Hero diverged)]
          | Summary                   | Avg      | TestHand Impact (Bet-Chk)           |
          | :------------------------ | :------: | :---------------------------------: |
          | Hero EV Adv               | -0.19    | **0.27** (-44.9)                    |
          | Hand EV                   | 0.41     | **0.34** (+30.9)                    |
          | Vuln - Range Avg          | 0.00     | **-0.15** (+15.6)                   |
          | Hand Blocker              | 3.62     | **1.00** (+1.7)                     |
          | Hand EQ                   | 0.49     | **0.24** (-0.5)                     |
        +-- NODE #6
              [Path]  TestHand [8d7d] -> P2->Check  (Reach: 50%->42%) | R: [0:28 1:28 2:42]
              [Summary Impact Analysis - Node #6 (Analysis Focus: TestHand - Hero diverged)]
              | Summary                   | Avg      | TestHand Impact (Bet-Chk)           |
              | :------------------------ | :------: | :---------------------------------: |
              | Hand EV                   | 0.41     | **0.34** (+54.6)                    |
              | Hero EQ Adv               | -0.03    | **0.17** (-35.6)                    |
              | Phys: Kicker              | 0.65     | **0.50** (+7.9)                     |
              | Bluff Efficiency          | 0.61     | **0.63** (+6.8)                     |
              | EQ - Nut Range Avg        | 0.27     | **-0.01** (+4.4)                    |
            +-- NODE #21
                  [Path]  TestHand [8d7d] -> P2->Check  (Reach: 21%->97%) | R: [0:03 1:00 2:96]
                  [Summary Impact Analysis - Node #21 (Analysis Focus: TestHand - Hero diverged)]
                  | Summary                   | Avg      | TestHand Impact (Bet-Chk)           |
                  | :------------------------ | :------: | :---------------------------------: |
                  | EQ - Nut MDF              | 0.25     | **0.00** (+6.5)                     |
                  | Regret: Fold-Call         | 0.89     | **1.43** (-6.5)                     |
                  | Phys: Kicker              | 0.65     | **0.50** (+3.8)                     |
                  | Vuln - Range Avg          | 0.00     | **-0.15** (+3.6)                    |
                  | OOP Rng EV                | 0.59     | **0.36** (-2.3)                     |
                +-- LEAF #66 (Static Prototype)
                      -> Strat: [Check:99%] | Samples: [TestHand:8d7d] (Reach:20.3%)

## 3. Key Summary Drivers (Path-Dependent)

### **(Depth 0)** - Top 5 Drivers:
- **Rank in Range (EQ)**: 0.03 (Avg: 0.50) | Impact: +30.35
- **Hand EQ**: 0.24 (Avg: 0.49) | Impact: +17.70
- **MDF (Decision)**: -1.00 (Avg: 0.65) | Impact: -6.30
- **Phys: Kicker**: 0.50 (Avg: 0.65) | Impact: -3.15
- **Hand EV**: 0.34 (Avg: 0.41) | Impact: -0.40

### **(Depth 1)** - Top 4 Drivers:
- **Hero EV Adv**: 0.27 (Avg: -0.19) | Impact: -44.88
- **Hand EV**: 0.34 (Avg: 0.41) | Impact: +30.94
- **Vuln - Range Avg**: -0.15 (Avg: 0.00) | Impact: +15.60
- **Hand Blocker**: 1.00 (Avg: 3.62) | Impact: +1.67

### **(Depth 2)** - Top 3 Drivers:
- **Hand EV**: 0.34 (Avg: 0.41) | Impact: +54.60
- **Hero EQ Adv**: 0.17 (Avg: -0.03) | Impact: -35.60
- **Phys: Kicker**: 0.50 (Avg: 0.65) | Impact: +7.90

### **(Depth 3)** - Top 2 Drivers:
- **EQ - Nut MDF**: 0.00 (Avg: 0.25) | Impact: +6.50
- **Regret: Fold-Call**: 1.43 (Avg: 0.89) | Impact: -6.47

\end{promptlisting}

LLM Response 3:
\begin{promptlisting}
## 1. Strategy Summary

The model applies a polarized bluffing strategy where "Pure Air" (7c6c, 16% EQ) mostly gives up but occasionally bluffs, whereas "Semi-Bluffs" (8d7d, 24% EQ) bet more frequently (40-50%) to realize their higher equity and draw potential.

## 2. Path Logic & Divergence

The decision tree separates hands based on their potential to improve (Equity) and their utility as bluffs (Blockers/EV).

* **Node #0 (The "Worthlessness" Filter):**
* **Logic:** The router isolates hands with very low **Rank in Range** (<0.05) and negative **MDF** metrics. Both Hero (7c6c) and Shadows (87/76) are identified as the "bottom of range," filtering out all value hands.
* **Result:** All low-equity hands proceed to Node #1.


* **Node #1 (The "Draw vs. Air" Split):**
* **Logic:** This is the critical divergence point. The router uses **Hand EV** and **Hand EQ** to distinguish "Pure Trash" from "Draws."
* **Hero (7c6c):** Has significantly lower **Hand EQ** (0.16) and **Hand EV** (0.29). It is routed primarily to **Path 0** (Node #15), treated as a "Pure Bluff" candidate that mostly checks but takes a high-frequency betting line (Leaf #46) when it *does* bluff to maximize fold equity.
* **Shadows (8s7s, etc.):** Have higher **Hand EQ** (0.24) due to gutshot/backdoor potential. They are routed to **Path 2** (Node #6), entering a "Semi-Bluff" line.


* **Theory:** The model adheres to the GTO principle that *zero-equity hands* (Hero) are often indifferent (mix check/bet or check/fold), while *low-equity draws* (Shadows) are incentivized to bet aggressively to realize equity or fold out better hands.


* **Node #6 & #21 (The Semi-Bluff Mix):**
* **Logic:** The Shadows (and the Test Case) traverse these nodes. Summaries such as **Bluff Efficiency** (+6.8 impact) and **Regret: Fold-Call** drive the decision.
* **Result:** Shadows reach a mix of checking (Leaf #66) and betting (inferred branch), resulting in a higher overall betting frequency (~45-55%) compared to Hero (~29%).



## 3. Test Case Solution

* **Predicted Mixed Strategy:** **Check: ~54% / Bet: ~46%** (Aggressive Semi-Bluff Mix)
* **Reasoning:**
* **Summary Matching:** The Test Case hand (8d7d) has **Hand EQ (0.24)** and **Hand EV (0.34)**, which are close to the continuing shadow hands (8s7s, 8h7h) and distinct from Hero (0.16 / 0.29).
* **Strategic Classification:** Unlike Hero (7c6c), which is "Pure Air," 8d7d holds a gutshot straight draw (8-7 on Q-J-5 requires 9-T).
* **Path Prediction:** Because its "Depth 1" drivers (Hand EV, Vuln) match the Shadows perfectly, 8d7d will be routed away from Hero's "Pure Bluff" path at Node #1 and into the "Semi-Bluff" path (Node #6).
* **Outcome:** It will adopt the Shadow strategy profile: a balanced mix favoring aggression more than Hero, utilizing its equity to semi-bluff.
\end{promptlisting}

\end{document}